\documentclass{article}

\usepackage{microtype}
\usepackage{graphicx}
\usepackage{subfigure}
\usepackage{booktabs} 

\usepackage{hyperref}

\usepackage[accepted]{icml2024}

\usepackage{amsmath}
\usepackage{amssymb}
\usepackage{mathtools}
\usepackage{amsthm}

\usepackage[capitalize,noabbrev]{cleveref}

\usepackage{bbm}
\usepackage{multirow}
\usepackage{nicefrac}

\theoremstyle{plain}
\newtheorem{theorem}{Theorem}[section]
\newtheorem{proposition}[theorem]{Proposition}
\newtheorem{lemma}[theorem]{Lemma}

\theoremstyle{definition}

\theoremstyle{remark}
\newtheorem{remark}[theorem]{Remark}

\DeclareMathOperator*{\argmin}{arg\,min}

\renewcommand{\P}{\mathbbm{P}}
\newcommand{\E}{\mathbbm{E}}
\newcommand{\R}{\mathbb{R}}

\makeatletter
\DeclareRobustCommand{\cev}[1]{%
  {\mathpalette\do@cev{#1}}%
}
\newcommand{\do@cev}[2]{%
  \vbox{\offinterlineskip
    \sbox\z@{$\m@th#1 x$}%
    \ialign{##\cr
      \hidewidth\reflectbox{$\m@th#1\vec{}\mkern4mu$}\hidewidth\cr
      \noalign{\kern-\ht\z@}
      $\m@th#1#2$\cr
    }%
  }%
}
\makeatother

\makeatletter
\newcommand{\subsetcong}{\mathrel{\mathpalette\subset@cong\relax}}

\newcommand{\subset@cong}[2]{%
  \vbox{\offinterlineskip\m@th
    \ialign{\hfil$#1##$\hfil\cr
      \sim\cr\subset\cr
    }%
  }%
}

\icmltitlerunning{Bridging discrete and continuous state space: Exploring the Ehrenfest process in time-continuous diffusion models}

\begin{document}

\twocolumn[
\icmltitle{Bridging discrete and continuous state spaces:\\Exploring the Ehrenfest process in time-continuous diffusion models}

\icmlsetsymbol{equal}{*}

\begin{icmlauthorlist}
\icmlauthor{Ludwig Winkler}{equal,TU}
\icmlauthor{Lorenz Richter}{equal,ZIB,dida}
\icmlauthor{Manfred Opper}{TU,Birm,Pots}

\end{icmlauthorlist}

\icmlaffiliation{TU}{Technical University of Berlin}
\icmlaffiliation{ZIB}{Zuse Institute Berlin}
\icmlaffiliation{dida}{dida Datenschmiede GmbH}
\icmlaffiliation{Birm}{University of Birmingham}
\icmlaffiliation{Pots}{University of Potsdam}

\icmlcorrespondingauthor{Ludwig Winkler}{winkler@tu-berlin.de}
\icmlcorrespondingauthor{Lorenz Richter}{richter@zib.de}

\vskip 0.3in]

\printAffiliationsAndNotice{\icmlEqualContribution} 

\begin{abstract}
Generative modeling via stochastic processes has led to remarkable empirical results as well as to recent advances in their theoretical understanding. In principle, both space and time of the processes can be discrete or continuous. In this work, we study time-continuous Markov jump processes on discrete state spaces and investigate their correspondence to state-continuous diffusion processes given by SDEs. 
In particular, we revisit the \textit{Ehrenfest process}, which converges to an Ornstein-Uhlenbeck process in the infinite state space limit. Likewise, we can show that the time-reversal of the Ehrenfest process converges to the time-reversed Ornstein-Uhlenbeck process. This observation bridges discrete and continuous state spaces and allows to carry over methods from one to the respective other setting. Additionally, we suggest an algorithm for training the time-reversal of Markov jump processes which relies on conditional expectations and can thus be directly related to denoising score matching. We demonstrate our methods in multiple convincing numerical experiments. 
\end{abstract}

\section{Introduction}
    Generative modeling based on stochastic processes has led to state-of-the-art performance in multiple tasks of interest, all aiming to sample artificial data from a distribution that is only specified by a finite set of training data \cite{nichol2021improved}. The general idea is based on the concept of time-reversal: we let the data points \textit{diffuse} until they are close to the equilibrium distribution of the process, from which we assume to be able to sample readily, such that the time-reversal then brings us back to the desired target distribution \cite{sohl2015deep}. In this general setup, one can make several choices and take different perspectives. While the original attempt considers discrete-time, continuous-space processes \cite{ho2020denoising}, one can show that in the small step-size limit the models converge to continuous-time, continuous-space processes given by stochastic differential equations (SDEs) \cite{song2020score}. This continuous time framework then allows fruitful connections to mathematical tools such as partial differential equations, path space measures and optimal control \cite{berner2022optimal}. As an alternative, one can consider discrete state spaces in continuous time via Markov jump processes, which have been suggested for generative modeling in \citet{campbell2022continuous}. Those are particularly promising for problems that naturally operate on discrete data, such as, e.g., text, images, graph structures or certain biological data, to name just a few. While discrete in space, an appealing property of those models is that time-discretization is not necessary -- neither during training nor during inference\footnote{Note that this is not true for the time- and space-continuous SDE case, where training can be done simulation-free, however, inference relies on a discretization of the reverse stochastic process. However, see \Cref{sec: tau leaping} for high-dimensional settings in Markov jump processes.}.

    While the connections between Markov jump processes and state-continuous diffusion processes have been studied extensively (see, e.g., \citet{kurtz1972relationship}), a relationship between their time-reversals has only been looked at recently, where an exact correspondence is still elusive \cite{santos2023blackout}. In this work, we make this correspondence more precise, thus bridging the gap between discrete-state generative modeling with Markov jump processes and the celebrated continuous-state score-based generative modeling. A key ingredient will be the so-called \textit{Ehrenfest process}, which can be seen as the discrete-state analog of the Ornstein-Uhlenbeck process, that is usually employed in the continuous setting, as well as a new loss function that directly translates learning rate functions of a time-reversed Markov jump process to score functions in the continuous-state analog. Our contributions can be summarized as follows:
\begin{itemize}
    \item We propose a loss function via conditional expectations for training state-discrete diffusion models, which exhibits advantages compared to previous loss functions.
    \item We introduce the \textit{Ehrenfest process} and derive the jump moments of its time-reversed version.
    \item Those jump moments allow an exact correspondence to score-based generative modeling, such that, for the first time, the two methods can now be directly linked to one another.
    \item In consequence, the bridge between discrete and continuous state space brings the potential that one setting can benefit from the respective other.
\end{itemize}

This paper is organized as follows. After listing related work in \Cref{sec: related work} and defining notation in \Cref{sec: notation}, we introduce the time-reversal of Markov jump processes in \Cref{sec: time-reversed jump processes} and propose a loss function for learning this reversal in \Cref{sec: loss functions}. We define the Ehrenfest process in \Cref{sec: Ehrenfest process}  and study its convergence to an SDE in \Cref{sec: scaled Ehrenfest process}. In \Cref{sec: Ehrenfest and score} we then establish the connection between the time-reversed Ehrenfest process and score-based generative modeling. \Cref{sec: computational aspects} is devoted to computational aspects and \Cref{sec: experiments} provides some numerical experiments that demonstrate our theory. Finally, we conclude in \Cref{sec: conclusion}.

\subsection{Related work}
\label{sec: related work}

Starting with a paper by \citet{sohl2015deep}, a number of works have contributed to the success of diffusion-based generative modeling, all in the continuous-state setting, see, e.g.,~\citet{ho2020denoising,song2020improved,kingma2021variational,nichol2021improved,vahdat2021score}. We shall highlight the work by \citet{song2020score}, which derives an SDE formulation of score-based generative modeling and thus builds the foundation for further theoretical developments \cite{berner2022optimal,richter2023improved}. We note that the underlying idea of time-reversing a diffusion process dates back to work by \citet{nelson1967dynamical,anderson1982reverse}.

Diffusion models on discrete state spaces have been considered by \citet{hoogeboom2021argmax} based on appropriate binning operations of continuous models. \citet{song2020denoising} proposed a method for discrete categorical data, however, did not perform any experiment. A purely discrete diffusion model, both in time and space, termed \textit{Discrete Denoising Diffusion Probabilistic Models (D3PMs)} has been introduced in \citet{austin2021structured}. Continuous-time Markov jump processes on discrete spaces have first been applied to generative modeling in \citet{campbell2022continuous}, where, however, different forward processes have been considered, for which the forward transition probability is approximated by solving the forward Kolmogorov equation. \citet{sun2022score} introduced the idea of categorical ratio matching for continuous-time Markov Chains by learning the conditional distribution occurring in the transition ratios of the marginals when computing the reverse rates. Recently, in a similar setting, \citet{santos2023blackout} introduced a pure death process as the forward process, for which one can derive an alternative loss function. Further, they formally investigate the correspondence between Markov jump processes and SDEs, however, in contrast to our work, without identifying a direct relationship between the corresponding learned models.

Finally, we refer to the monographs \citet{gardiner1985handbook, van1992stochastic, bremaud2013markov} for a general introduction to Markov jump processes.

\subsection{Notation}
\label{sec: notation}
For transition probabilities of a Markov jump process $M$ we write $p_{t|s}(x | y):=\mathbb{P}\left(M(t)=x | M(s)=y \right)$ for $s, t\in [0, T]$ and $x,y\in \Omega$. With $p_t(x)$ we denote the (unconditional) probability of the process at time $t$. We use $p_\mathrm{data} := p_0$. With $\delta_{x, y}$ we denote the Kronecker delta. For a function $f$, we say that $f(x) \in o(g(x))$ if $\lim_{x\to 0} \frac{f(x)}{g(x)} = 0$.

\section{Time-reversed Markov jump processes}
\label{sec: time-reversed jump processes}

We consider Markov jump processes $M(t)$ that run on the time interval $[0, T] \subset \R$ and are allowed to take values in a discrete set $\Omega \subsetcong \mathbb{Z}^d$. Usually, we consider $\Omega \cong \left\{0, \dots, S \right\}^d$ such that the cardinality of our space is $|\Omega| = (S+1)^d$. Jumps between the discrete states appear randomly, where the rate of jumping from state $y$ to $x$ at time $t$ is specified by the function $r_t(x | y)$. The jump rates determine the jump probability in a time increment $\Delta t$ via the relation
\begin{equation}
\label{eq: transition Markov jump}
    p_{t+\Delta t| t}(x | y) = \delta_{x, y} + r_t(x| y) \Delta t + o(\Delta t),
\end{equation}
i.e. the higher the rate and the longer the time increment, the more likely is a transition between two corresponding states. For a more detailed introduction to Markov jump processes, we refer to \Cref{sec: intro Markov jump processes}. In order to simulate the process backwards in time, we are interested in the rates of the time-reversed process $\cev{M}(t)$, which determine the backward transition probability via
\begin{equation}
    p_{t-\Delta t| t}(x | y) = \delta_{x, y} + \cev{r}_t(x| y) \Delta t + o(\Delta t).
\end{equation}
The following lemma provides a formula for the rates of the time-reversed process, cf. \citet{campbell2022continuous}.

\begin{lemma}
\label{lem: backward rates}
For two states $x, y \in \Omega$, the transition rates of the time-reversed process $\cev{M}(t)$ are given by 
    \begin{equation}
    \label{eq: backward rates}
        \cev{r}_t(y | x) = \E_{x_0 \sim p_{0|t}(x_0 | x)}\left[ \frac{p_{t|0}(y | x_0)}{p_{t|0}(x | x_0)} \right] r_t(x | y),
    \end{equation}
where $r_t$ is the rate of the forward process $M(t)$.
\end{lemma}
\begin{proof}
    See \Cref{app: proofs}.
\end{proof}

\begin{remark}[Conditional expectation]
    We note that the expectation appearing in \eqref{eq: backward rates} is a conditional expectation, conditioned on the value $M(t) = x$. This can be compared to the SDE setting, where the time-reversal via the score function can also be written as a conditional expectation, namely $\nabla_x \log p_t^\mathrm{SDE}(x) = \E_{x_0 \sim p_{0|t}^\mathrm{SDE}(x_0 | x)}\left[\nabla_x \log p_{t | 0}^\mathrm{SDE}(x |x_0) \right]$, see \Cref{lem: score as conditional expectation} in the appendix for more details. We will elaborate on this correspondence in \Cref{sec: Ehrenfest and score}.
\end{remark}

While the forward transition probability $p_{t|0}$ can usually be approximated (e.g. by solving the corresponding master equation, see \Cref{sec: intro Markov jump processes}), the time-reversed transition function $p_{0|t}$ is typically not tractable, and we therefore must resort to a learning task. One idea is to approximate $p_{0|t} \approx p^\theta_{0|t}$ by a distribution parameterized in $\theta \in \R^p$ (e.g. via neural networks), see, e.g. \citet{campbell2022continuous} and \Cref{sec: Learning the reversed transition probability}. We suggest an alternative method in the following.

\subsection{Loss functions via conditional expectations}
\label{sec: loss functions}

Recalling that any conditional expectation can be written as an $L^2$ projection (see \Cref{lem: minimizer L2 error} in the appendix), we define the loss
\begin{equation}
\label{eq: cond exp loss}
    \mathcal{L}_y(\varphi_y) = \E\left[\left(\varphi_y(x, t) - \frac{ p_{t|0}(y|x_0)}{ p_{t|0}(x | x_0)} \right)^2 \right],
\end{equation}
where the expectation is over $x_0 \sim p_\mathrm{data}, t\sim\mathcal{U}(0,T),x\sim p_{t|0}(x|x_0)$. Assuming a sufficiently rich function class $\mathcal{F}$, it then holds that the minimizer of the loss equals the conditional expectation in \Cref{lem: backward rates} for any $y \in \Omega$, i.e.
\begin{align}
\begin{split}
   \argmin_{\varphi_y \in \mathcal{F}} \mathcal{L}_y(\varphi_y) = \mathbb{E}_{x_0 \sim p_{0|t}(x_0 | x)} &\left[ \frac{ p_{t|0}(y|x_0)}{ p_{t|0}(x | x_0)} \right].
\end{split}
\end{align}

We can thus directly learn the conditional expectation. In contrast to approximating the reverse transition probability $p_{0|t}$, this has the advantage that we do not need to model a distribution, but a function, which is less challenging from a numerical perspective. Furthermore, we will see that the conditional expectation can be directly linked to the score function in the SDE setting, such that our approximating functions $\varphi_y$ can be directly linked to the approximated score. We note that the loss has already been derived in a more general version in \citet{meng2022concrete} and applied to the setting of Markov jump processes in \citet{lou2023discrete}, however, following a different derivation. A potential disadvantage of the loss \eqref{eq: cond exp loss}, on the other hand, is that we may need to approximate different functions $\varphi_y$ for different $y \in \Omega$. This, however, can be coped with in two ways. On the one hand, we may focus on \textit{birth-death processes}, for which $r(y | x)$ is non-zero only for $y = x \pm 1$, such that we only need to learn $2$ instead of $S-1$ functions $\varphi_y$. In the next section we will argue that birth-death process are in fact favorable for multiple reasons. On the one hand, we can do a Taylor expansion such that for certain processes it suffices to only consider one approximating function, as will be shown in \Cref{rem: Taylor expanstion of ratio}. 

\section{The Ehrenfest process}
\label{sec: Ehrenfest process}

In principle, we are free to choose any forward process $M(t)$ for which we can compute the forward transition probabilities $p_{t|0}$ and which is close to its stationary distribution after a not too long run time $T$. In the sequel, we argue that the \textit{Ehrenfest process} is particularly suitable -- both from a theoretical and practical perspective. For notational convenience, we make the argument in dimension $d=1$, noting, however, that a multidimensional extension is straightforward. For computational aspects in high-dimensional spaces we refer to \Cref{sec: modeling of dimensions}.

We define the Ehrenfest process\footnote{The Ehrenfest process was introduced by the Russian-Dutch and German physicists Tatiana and Paul Ehrenfest to explain the second law of thermodynamics, see \citet{ehrenfest1907zwei}.} as
\begin{equation}
\label{eq: definition Ehrenfest}
    E_S(t) := \sum_{i=1}^{S} Z_i(t),
\end{equation}
where each $Z_i$ is a process on the state space $\Omega = \{0, 1\}$ with transition rates $r(0 | 1) = r(1 | 0) = \frac{1}{2}$ (sometimes called \textit{telegraph} or \textit{Kac process}). We note that the \textit{Ehrenfest process} is a birth-death process with values in $\{0, \dots, S\}$ and transition rates
\begin{equation}
\label{eq: Ehrenfest rates}
    r(x+1|x) = \frac{1}{2}(S-x), \qquad r(x-1|x) = \frac{x}{2}.
\end{equation}

We observe that we can readily transform the time-independent rates in \eqref{eq: Ehrenfest rates} to time-dependent rates
\begin{equation}
\label{eq: rates with time transformation}
    r_t(x \pm 1 | x) := \lambda_t \, r(x\pm1|x)
\end{equation}
via a time transformation, where $\lambda:[0,T] \to \R$, see \Cref{app: time transformation}. Without loss of generality, we will focus on the time-independent rates \eqref{eq: Ehrenfest rates} in the sequel.

One compelling property of the Ehrenfest process is that we can sample without needing to simulate trajectories.

\begin{lemma}
\label{lem: forward transition prob Ehrenfest}
Assuming $E_S(0) = x_0$, the Ehrenfest process can be written as
\begin{equation}
    E_S(t) = E_{0,S}(t) + E_{1,S}(t),
\end{equation}
where $E_{0, S}(t) \sim B(S - x_0, 1 - f(t))$ and $E_{1, S}(t) \sim B(x_0, f(t))$ are independent binomial random variables and $f(t) := 
 \frac{1}{2}\left(1 + e^{-t}\right)$. Consequently, the forward transition probability is given by the discrete convolution
     \begin{equation}
     \label{eq: convolution of binomials}
     p_{t|0}(x | x_0)  
     = \sum_{z\in \Omega}  \P\left(E_{0,S}(t) = z \right) \P\left(E_{1,S}(t) = x-z \right).
 \end{equation}
 
\end{lemma}
\begin{proof}
    See \Cref{app: proofs}.
\end{proof}

We note that the sum in \eqref{eq: convolution of binomials} can usually be numerically evaluated without great effort.

\subsection{Convergence properties in the infinite state space limit}
\label{sec: scaled Ehrenfest process}

It is known that certain (appropriately scaled) Markov jump processes converge to state-continuous diffusion processes when the state space size $S + 1$ tends to infinity (see, e.g., \citet{kurtz1972relationship, gardiner1985handbook}). For the Ehrenfest process, this convergence can be studied quite rigorously. To this end, let us introduce the scaled Ehrenfest process
\begin{equation}
\label{eq: scaled Ehrenfest}
    \widetilde{E}_S(t) := \frac{2}{\sqrt{S}}\left(E_{S}(t) -  \frac{S}{2}\right)
\end{equation}
with transition rates

\begin{align}
\label{eq: scaled Ehrenfest rates}
    r\left(x\pm \frac{2}{\sqrt{S}}\bigg|x\right) &= \frac{\sqrt{S}}{4}(\sqrt{S} \mp x),
\end{align}

now having values in $\Omega = \left\{-\sqrt{S},-\sqrt{S}+\frac{2}{\sqrt{S}}, \dots, \sqrt{S} \right\}$. We are interested in the large state space limit $S \to \infty$, noting that this implies $\frac{2}{\sqrt{S}} \to 0$ for the transition steps, thus leading to a refinement of the state space. The following convergence result is shown in \citet[Theorem 4.1]{sumita2004numerical}.

\begin{proposition}[State space limit of Ehrenfest process]
\label{prop: convergence of forward Ehrenfest}
    In the limit $S \to \infty$, the scaled Ehrenfest process $\widetilde{E}_S(t)$ converges in law to the Ornstein-Uhlenbeck process $X_t$ for any $t \in [0, T]$, where $X_t$ is defined via the SDE
    \begin{equation}
    \label{eq: OU process}
        \mathrm d X_t = -X_t \, \mathrm d t + \sqrt{2} \, \mathrm d W_t,
    \end{equation}
    with $W_t$ being standard Brownian motion.
\end{proposition}

For an illustration of the convergence we refer to \Cref{fig: reverse OU and Ehrenfest}.

Note that the convergence of the scaled Ehrenfest process to the Ornstein-Uhlenbeck process implies
\begin{equation}
\label{eq: approx of forward transition by Gaussian}
    p_{t|0}(x|x_0) \approx p_{t|0}^{\mathrm{OU}}(x|x_0) := \mathcal{N}(x; \mu_t(x_0), \sigma_t^2)
\end{equation}
with $\mu_t(x_0) = x_0 e^{-t}$ and $\sigma_t^2 = (1 - e^{-2t})$. For the quantity in the conditional expectation \eqref{eq: backward rates} we can thus compute
\begin{subequations}
\begin{align}
\label{eq: Gaussian approximation of ratio}
    &\frac{p_{t|0}\left(x \pm \delta \Big| x_0\right)}{p_{t|0}(x|x_0)} \approx \exp\left(\frac{\mp 2(x - \mu_t(x_0))\delta - \delta^2}{2 \sigma^2_t}\right) \\
    \label{eq: Taylor of fraction}
    &\hspace{-0.1cm}\approx \exp\hspace{-0.1cm}\left( \hspace{-0.1cm}-\frac{\delta^2}{2 \sigma_t^2}\right)\hspace{-0.15cm}\left(\hspace{-0.05cm}1 \mp \frac{(x-\mu_t(x_0))\delta}{\sigma^2} + \frac{\left((x-\mu_t(x_0))\delta\right)^2}{2\sigma^4} \right),
\end{align}
\end{subequations}
where we used the shorthand $\delta := \frac{2}{\sqrt{S}}$.

\begin{remark}[Learning of conditional expectation]
\label{rem: Taylor expanstion of ratio}
    Note that the approximation \eqref{eq: Taylor of fraction} allows us to define the loss
    \begin{align}
    \label{eq: Gauss loss}
    \begin{split}
        &\mathcal{L}_\mathrm{Gau\ss}(\varphi) := \\
        &\quad\E\left[\left(\varphi(x,t) - \exp\left(\frac{\mp 2(x - \mu_t(x_0))\delta - \delta^2}{2 \sigma^2_t}\right)\right)^2\right].
    \end{split}
    \end{align}
    Further, we can write 
    \begin{align}
    \begin{split}
        &\mathbb{E}_{x_0}\left[\frac{p_{t|0}\left(x \pm \delta \Big| x_0\right)}{p_{t|0}(x|x_0)}\right] \approx \exp\left( -\frac{\delta^2}{2 \sigma_t^2}\right)\\
        &\left(1 \mp \frac{(x-\E_{x_0}\left[\mu_t(x_0)\right])\delta}{\sigma^2} + \frac{\E_{x_0}\left[\left((x-\mu_t(x_0))\delta\right)^2\right]}{2\sigma^4} \right),
    \end{split}
    \end{align}
    where $x_0 \sim p_{0|t}(x_0 | x)$. In consequence, this allows us to consider the loss functions
    \begin{equation}
    \label{eq: first Taylor ratio loss}
            \mathcal{L}_\mathrm{Taylor}(\varphi_1) := \E\left[\left(\varphi_1(x, t) - \mu_t(x_0) \right)^2 \right],
    \end{equation}
    and
    \begin{equation}
            \mathcal{L}_\mathrm{Taylor,2}(\varphi_2) := \E\left[\left(\varphi_2(x, t) - \left((x-\mu_t(x_0))\delta\right)^2 \right)^2 \right],
    \end{equation}
    where the expectations are over $x_0 \sim p_\mathrm{data}, t\sim\mathcal{U}(0,T),x\sim p_{t|0}(x|x_0)$. We can also only consider the first order term in the Taylor expansion \eqref{eq: Taylor of fraction}, such that we then only have to approximate one instead of two functions.
\end{remark}

Since the scaled forward Ehrenfest process converges to the Ornstein-Uhlenbeck process, we can expect the time-reversed scaled Ehrenfest process to converge to the time-reversal of the Ornstein-Uhlenbeck process. We shall study this conjecture in more detail in the sequel.

\subsection{Connections between time-reversal of Markov jump processes and score-based generative modeling}
\label{sec: Ehrenfest and score}

Inspecting \Cref{lem: backward rates}, which specifies the rate function of a backward Markov jump process, we realize that the time-reversal essentially depends on two things, namely the forward rate function with switched arguments as well as the conditional expectation of the ratio between two forward transition probabilities. To gain some intuition, let us first assume that the state space size $S + 1$ is large enough and that the transition density $p_{t | 0}$ can be extended to $\R$ (which we call $\overline{p}_{t | 0}$) such that it can be approximated via a Taylor expansion. We can then assume that
\begin{equation}
    r\left(x \pm \frac{2}{\sqrt{S}}\bigg| x\right) \approx  r\left(x \bigg| x \mp \frac{2}{\sqrt{S}} \right)
\end{equation}
as well as
\begin{subequations}
\begin{align}
\label{eq: Taylor fraction continous extension}
    \frac{p_{t|0}\left(x \pm \frac{2}{\sqrt{S}}\Big| x_0\right)}{p_{t|0}(x|x_0)} &\approx \frac{\overline{p}_{t|0}(x|x_0) \pm \frac{2}{\sqrt{S}} \nabla \overline{p}_{t|0}(x|x_0)}{\overline{p}_{t|0}(x|x_0)}\\
    &= 1 \pm \frac{2}{\sqrt{S}} \nabla \log \overline{p}_{t| 0}(x | x_0),
\end{align}
\end{subequations}

where the conditional expectation of $\nabla \log \overline{p}_{t| 0}(x | x_0)$ is reminiscent of the score function in SDE-based diffusion models (cf. \Cref{lem: score as conditional expectation} in the appendix). This already hints at a close connection between the time-reversal of Markov jump processes and score-based generative modeling. Further, note that \eqref{eq: Taylor fraction continous extension} corresponds to \eqref{eq: Taylor of fraction} for large enough $S$ and $p_{t| 0} \approx \overline{p}_{t|0}^\mathrm{OU}$.

We shall make the above observation more precise in the following. To this end, let us study the first and second jump moments of the Markov jump process, given as
\begin{align}
       b(x) &= \sum_{y \in \Omega, y \neq x} (y-x) r(y | x), \\
    D(x) &= \sum_{y \in \Omega, y \neq x} (y-x)^2 r(y | x),
\end{align}
see \Cref{app: convergence of jump processes}. For the scaled Ehrenfest process \eqref{eq: scaled Ehrenfest} we can readily compute
\begin{equation}
    b(x) = -x, \qquad D(x) = 2,
\end{equation}
which align with the drift and diffusion coefficient (which is the square root of $D$) of the Ornstein-Uhlebeck process in \Cref{prop: convergence of forward Ehrenfest}. In particular, we can show the following relation between the jump moments of the forward and the backward Ehrenfest processes, respectively.

\begin{proposition}
\label{prop: jump moment reversed Ehrenfest}
    Let $b$ and $D$ be the first and second jump moments of the scaled Ehrenfest process $\widetilde{E}_S$. The first and second jump moments of the time-reversed scaled Ehrenfest $\cev{\widetilde{E}}_S$ are then given by
    \begin{align}
    \begin{split}
    \label{eq: reversed first jump moment}
        \cev{b}(x, t) &= -b(x) + D(x) \, \E_{x_0 \sim p_{0|t}(x_0 | x)}\left[ \frac{\Delta_S p_{t | 0}(x|x_0)}{p_{t | 0}(x|x_0)} \right] \\
        &\qquad + o(S^{-1/2}),\end{split}\\
        \label{eq: reversed second jump moment}\cev{D}(x) &= D(x) + o(S^{-1/2}),
    \end{align}
    where 
    \begin{equation}
        \Delta_S p_{t | 0}(x|x_0) := \frac{p_{t | 0}(x + \frac{2}{\sqrt{{S}}}|x_0) - p_{t | 0}(x|x_0)}{\frac{2}{\sqrt{S}}}
    \end{equation}
    is a one step difference and $p_{t|0}$ and $p_{0|t}$ are the forward and reverse transition probabilities of the scaled Ehrenfest process.
\end{proposition}

\begin{proof}
    See \Cref{app: proofs}.
\end{proof}

\begin{remark}[Convergence of the time-reversed Ehrenfest process]
\label{rem: convergence of time-reversed Ehrenfest}
    We note that \Cref{prop: jump moment reversed Ehrenfest} implies that the time-reversed Ehrenfest process in expected to converge in law to the time-reversed Ornstein-Uhlenbeck process. This can be seen as follows. For $S \to \infty$, we know via \Cref{prop: convergence of forward Ehrenfest} that the forward Ehrenfest process converges to the Ornstein-Uhlenbeck process, i.e. $p_{t|0}$ converges to $p_{t|0}^\mathrm{OU}$, where $p_{t|0}^\mathrm{OU}(x | x_0)$ is the transition  density of the Ornstein-Uhlenbeck process \eqref{eq: OU process} starting at $X_0 = x_0$. Together with the fact that the finite difference approximation operator $\Delta_S$ converges to the first derivative, this implies that $\E_{x_0 \sim p_{0|t}(x_0 | x)}\left[ \frac{\Delta_S p_{t | 0}(x|x_0)}{p_{t | 0}(x|x_0)} \right]$ is expected to converge to $\E_{x_0 \sim p_{0|t}^\mathrm{OU}(x_0 | x)}\left[ \nabla \log p_{t|0}^\mathrm{OU}(x | x_0)  \right]$. Now, \Cref{lem: score as conditional expectation} in the appendix shows that this conditional expectation is the score function of the Ornstein-Uhlenbeck process, i.e. $\nabla \log p_t^\mathrm{OU}(x) = E_{x_0 \sim p_{0|t}^\mathrm{OU}(x_0 | x)}\left[ \nabla \log p_{t|0}^\mathrm{OU}(x | x_0)  \right]$. Finally, we note that the first and second jump moments converge to the drift and the square of the diffusion coefficient of the limiting SDE, respectively \cite{gardiner1985handbook}.
    Therefore, the scaled time-reversed Ehrenfest process $\cev{\widetilde{E}}_S(t)$ is expected to converge in law to the process $Y_t$ given by
    \begin{equation}
    \label{eq: time-reversed OU}
    \mathrm d Y_t = \left(Y_t + 2 \nabla \log p_{T-t}^\mathrm{OU}(Y_t) \right) \mathrm d t + \sqrt{2} \, \mathrm d W_t,
\end{equation}
    which is the time-reversal of the Ornstein-Uhlenbeck process stated in \eqref{eq: OU process}. Note that we write \eqref{eq: time-reversed OU} as a forward process from $t = 0$ to $t=T$, where $W_t$ is a forward Brownian motion, which induces the time-transformation $t \mapsto T-t$ in the score function.
\end{remark}

\begin{remark}[Generalizations] 
Following the proof of \Cref{prop: jump moment reversed Ehrenfest}, we expect that the formulas for the first two jump moments of the time-reversed Markov jump process, stated in \eqref{eq: reversed first jump moment} and \eqref{eq: reversed second jump moment}, are valid for any (appropriately scaled) birth-death process whose transition rates fulfill
\begin{equation}
    \frac{1}{S}\left(r(x \pm \delta | x) - r(x | x\mp \delta)\right) = o(S^{-1}),
\end{equation}
where $\delta$ is a jump step size that decreases with the state space size $S + 1$.
\end{remark}

\begin{figure}
\centering
\includegraphics[width=\linewidth]{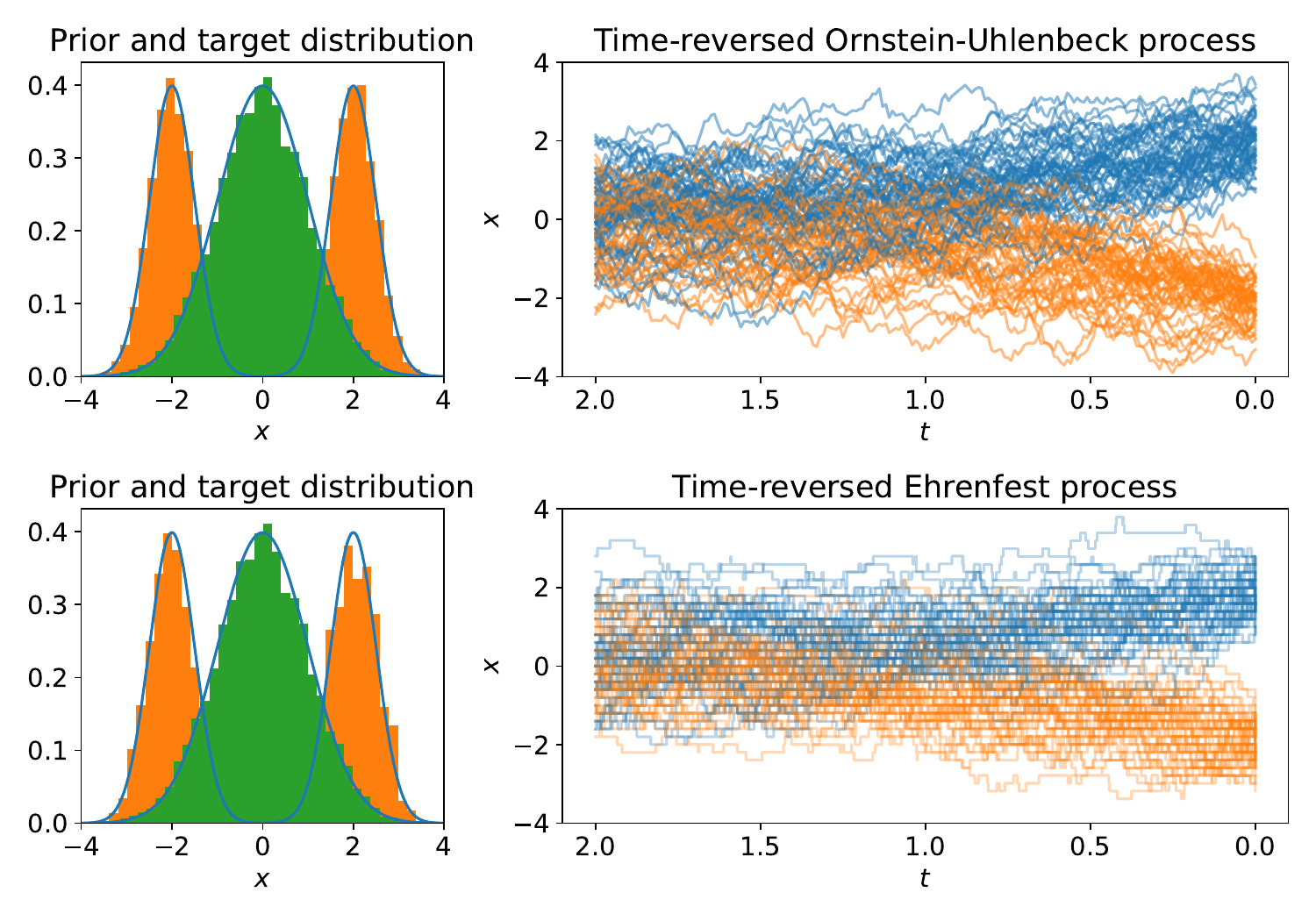}
\caption{We display two time-reversed processes from $t = 2$ to $t = 0$ that transport a standard Gaussian (left panels, in green) to a multimodal Gaussian mixture model (left panels, in orange), or a binomial distribution to a binomial mixture, respectively, once using a diffusion process in continuous space (upper panel) and once a time-reversed (scaled) Ehrenfest process in discrete space with $S=100$ (lower panel). Crucially, in both cases we use the (state-continuous) score function to employ the time-reversal, which for this problem is known analytically, see \Cref{app: Gaussian mixture score}. The plots demonstrate that the distributions of the processes seem indeed very close one another, implying that the approximation \eqref{eq: relation conditional expectation and score} is quite accurate even for a moderate state space size $S + 1$.}
\label{fig: reverse OU and Ehrenfest}
\end{figure}

Crucially, \Cref{rem: convergence of time-reversed Ehrenfest} shows that we can directly link approximations in the (scaled) state-discrete setting to standard state-continuous score-based generative modeling via
\begin{equation}
\label{eq: relation conditional expectation and score}
    \E_{x_0 \sim p_{0|t}(x_0 | x)}\hspace{-0.1cm}\left[ \frac{p_{t|0}(x \pm \frac{2}{\sqrt{S}} | x_0)}{p_{t|0}(x | x_0)} \right] \hspace{-0.1cm}\approx 1 \pm \frac{2}{\sqrt{S}} \nabla \log p_{t}^\mathrm{OU}(x),
\end{equation}

see also the proof of \Cref{prop: jump moment reversed Ehrenfest} in \Cref{app: proofs}. In particular, this allows for transfer learning between the two cases. E.g., we can train a discrete model and use the approximation of the conditional expectation (up to scaling) as the score function in a continuous model. Likewise, we can train a continuous model and approximate the conditional expectation by the score. We have illustrated the latter approach in \Cref{fig: reverse OU and Ehrenfest}, where we have used the (analytically available) score function that transports a standard Gaussian to a multimodal Gaussian mixture in a discrete-state Ehrenfest process that starts at a binomial distribution which is designed in such a way that it converges to the standard Gaussian for $S \to \infty$.

Similar to \eqref{eq: cond exp loss}, the correspondence \eqref{eq: relation conditional expectation and score} motivates to train a state-discrete scaled Ehrenfest model with the loss defined by
\begin{subequations}
\label{eq: forward OU loss}
\begin{align}
    \mathcal{L}_\mathrm{OU}(\widetilde{\varphi}) &:= \E\left[\left(\widetilde{\varphi}(x, t) - \nabla \log p_{t|0}^\mathrm{OU}(x|x_0)\right)^2\right] \\
    &=  \E\left[\left(\widetilde{\varphi}(x, t) +\frac{\left(x-\mu_t(x_0) \right)}{\sigma_t^2} \right)^2\right],
\end{align}
\end{subequations}
where the expectation is over $x_0 \sim p_\mathrm{data}, t\sim\mathcal{U}(0,T),x\sim p_{t|0}(x|x_0)$ and where $\mu_t(x_0) = x_0 e^{-t}$ and $\sigma_t^2 = (1 - e^{-2t})$, as before.
In fact, this loss is completely analog to the denoising score matching loss in the state-continuous setting.
We later set $\varphi = 1 \pm \frac{2}{\sqrt{S}}\widetilde{\varphi}^*$, where $\widetilde{\varphi}^*$ is the minimizer of \eqref{eq: forward OU loss}, to get the approximated conditional expectation.

\begin{remark}[Ehrenfest process as discrete-state DDPM]
\label{rem: Ehrenfest as discrete DDPM}
To make the above considerations more precise, note that we can directly link the discrete-space Ehrenfest process to pretrained score models in continuous space, such as, e.g., the celebrated \textit{denoising diffusion probabilistic models (DDPM)} \cite{ho2020denoising}. Those models usually transport a standard Gaussian to the target density that is supported on $[-1, 1]^d$. In order to cope with the fact that the scaled Ehrenfest process terminates (approximately) at a standard Gaussian irrespective of the size $S + 1$, we typically choose $S = 255^2$ such that the interval $[-1, 1]$ contains $256$ states that correspond to the RGB color values of images, recalling that the increments between the states are $\frac{2}{\sqrt{S}}$. Further, noting the actual Ornstein-Uhlenbeck process that DDPM is trained on, we employ the time scaling $\lambda_t = \frac{1}{2}\beta(t)$, where $\beta$ and further details are stated in \Cref{app: Ehrenfest - score based SDE}, and choose the (time-dependent) rates
\begin{equation}
        r_t\left(x\pm \frac{2}{\sqrt{S}}\bigg|x\right) = \beta(t) \frac{\sqrt{S}}{8}(\sqrt{S} \mp x),
\end{equation}
according to \eqref{eq: rates with time transformation} and \eqref{eq: scaled Ehrenfest rates}.

\end{remark}
 
\section{Computational aspects}
\label{sec: computational aspects}

In this section, we comment on computational aspects that are necessary for the training and simulation of the time-reversal of our (scaled) Ehrenfest process. For convenience, we refer to \Cref{alg: conditional expectation} and \Cref{alg: sampling} in \Cref{sec: computational aspects cond exp} for the corresponding training and sampling algorithms, respectively.

\subsection{Modeling of dimensions}
\label{sec: modeling of dimensions}

In order to make computations feasible in high-dimensional spaces $\Omega^d$, we typically factorize the forward process, such that each dimension propagates independently, cf. \citet{campbell2022continuous}. Note that this is analog to the Ornstein-Uhlenbeck process in score-based generative modeling, in which the dimensions also do not interact, see, e.g., \eqref{eq: OU process}. 

We thus consider 
\begin{equation}
    p_{t|0}(x|y) = \prod_{i=1}^d p_{t|0}^{(i)}(x^{(i)}|y^{(i)}),
\end{equation}
where $p^{(i)}_{t|0}$ is the transition probability for dimension $i \in \{1, \dots, d \}$ and $x^{(i)}$ is the $i$-th component of $x \in \Omega^d$. 

In \citet{campbell2022continuous} it is shown that the forward and backward rates then translate to
\begin{equation}
\label{eq: high dim forward rate}
    r_t(x | y) = \sum_{i=1}^d r_t^{(i)}(x^{(i)}|y^{(i)}) \Gamma_{x^{\neg i},y^{\neg i}},
\end{equation}
where $\Gamma_{x^{\neg i},y^{\neg i}}$ is one if all dimensions except the $i$-th dimension agree, and
\begin{equation}
\label{eq: high dim backward rate}
    \cev{r}_t(x | y) = \sum_{i=1}^d \E\left[ \frac{p_{t|0}(y^{(i)}|x_0^{(i)})}{p_{t|0}(x^{(i)}|x_0^{(i)})}\right] r_t^{(i)}(x^{(i)}|y^{(i)}) \Gamma_{x^{\neg i},y^{\neg i}},
\end{equation}
where the expectation is over $x_0^{(i)} \sim p_{0|t}(x^{(i)}_0 | x)$. Equation \eqref{eq: high dim backward rate} illustrates that the time-reversed process does not factorize in the dimensions even though the forward process does.

Note with \eqref{eq: high dim forward rate} that for a birth-death process a jump appears only in one dimension at a time, which implies that
\begin{equation}
r_t(x \pm \delta_i | x) = r_t^{(i)}(x^{(i)} \pm \delta^{(i)}_i|x^{(i)}),
\end{equation}
where now $\delta_i = (0, \dots, 0, \delta^{(i)}_i, 0, \dots, 0)^\top$ with $\delta^{(i)}_i$ being the jump step size in the $i$-th dimension. Likewise, \eqref{eq: high dim backward rate} becomes
\begin{equation}
\label{eq: high dim backward rate birth-death}
    \cev{r}_t(x\pm \delta_i | x) =  \E\left[ \frac{p_{t|0}(y^{(i)}|x_0^{(i)})}{p_{t|0}(x^{(i)}|x_0^{(i)})}\right] r_t^{(i)}(x^{(i)}|x^{(i)} + \delta_i^{(i)}),
\end{equation}
where the expectation is over $x_0^{(i)} \sim p_{0|t}(x^{(i)}_0 | x)$, which still depends on all dimensions. 

\begin{figure}[ht]
\centering
\includegraphics[width=\linewidth]{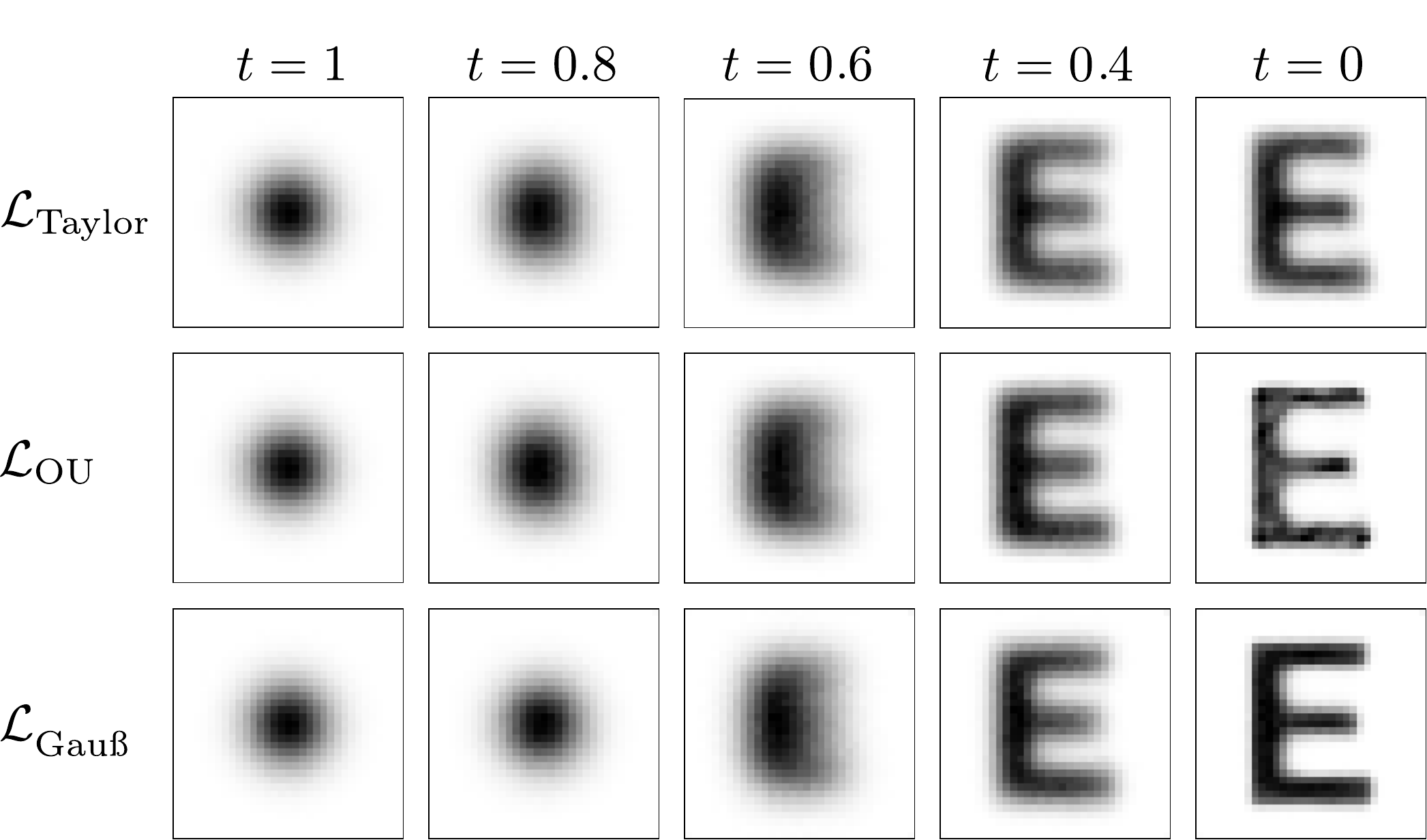}
\caption{We plot histograms of $500.000$ samples from the time-reversed scaled Ehrenfest process at different times. The processes have been trained with three different losses.}
\label{fig: toy example E}
\end{figure}

For each dimension $i \in \{1, \dots, d \}$ we can therefore approximate the conditional expectation appearing in \eqref{eq: high dim backward rate birth-death} via the loss function \eqref{eq: cond exp loss} with two functions $\varphi_{i,b} : \R^d \times [0, T] \to \R$ and $\varphi_{i,d}: \R^d \times [0, T] \to \R$. Alternatively, we can learn just two functions $\varphi_{b/d}: \R^d \times [0, T] \to \R^d$ for the entire space and identify $\varphi_{i,b/d} = \varphi_{b/d}^{(i)}$.

\subsection{$\tau$-leaping}
\label{sec: tau leaping}

The fact that jumps only happen in one dimension at a time implies that the naive implementation of changing component by component (e.g. by using the Gillespie’s algorithm, see \citet{gillespie1976general}) would require a very long sampling time. As suggested in \citet{campbell2022continuous}, we can therefore rely on $\tau$-leaping for an approximate simulation methods \cite{gillespie2001approximate}. The general idea is to not simulate jump by jump, but wait for a time interval of length $\tau$ and apply all jumps at once. One can show that the number of jumps is Poisson distributed with a mean of $\tau\, \cev{r}_t \,(x | y)$. For further details we refer to \Cref{alg: sampling}.

\section{Numerical experiments}
\label{sec: experiments}

In this section, we demonstrate our theoretical insights in numerical experiments. If not stated otherwise, we always consider the scaled Ehrenfest process defined in \eqref{eq: scaled Ehrenfest}. We will compare the different variants of the loss \eqref{eq: cond exp loss}, namely $\mathcal{L}_\mathrm{Gauss}$ defined in \eqref{eq: Gauss loss}, $\mathcal{L}_\mathrm{Taylor}$ defined in \eqref{eq: first Taylor ratio loss} and $\mathcal{L}_\mathrm{OU}$ defined in \eqref{eq: forward OU loss}.

\subsection{Illustrative example}

Let us first consider an illustrative example, for which the data distribution is tractable. We consider a process in $d=2$ with $S=32$, where the $(S+1)^d = 33^2$ different state combinations in $p_\mathrm{data}$ are defined to be proportional to the pixels of an image of the letter ``E''. Since the dimensionality is $d=2$, we can visually inspect the entire distribution at any time $t\in [0, T]$ by plotting 2-dimensional histograms of the simulated processes. With this experiment we can in particular check that modeling the dimensions of the forward process independently from one another (as explained in \Cref{sec: modeling of dimensions}) is no restriction for the backward process.
Indeed \Cref{fig: toy example E} shows that the time-reversed process, which is learned with (versions of) the loss \eqref{eq: cond exp loss}, can transport the prior distribution (which is approximately binomial, or, loosely speaking, a binned Gaussian) to the specified target. Again, note that this plot does not display single realizations, but entire distributions, which, in this case, are approximated with $500.000$ samples. We realize that in this simple problem $\mathcal{L}_\mathrm{Gau\ss}$ performs slightly better than $\mathcal{L}_\mathrm{OU}$ and $\mathcal{L}_\mathrm{Taylor}$. As expected, the approximations work sufficiently well even for a moderate state space size $S + 1$. As argued in \Cref{sec: scaled Ehrenfest process}, this should get even better with growing $S$. For further details, we refer to \Cref{app: illustrative example}.

\subsection{MNIST}

\begin{figure}[H]
\centering
\includegraphics[width=\linewidth]{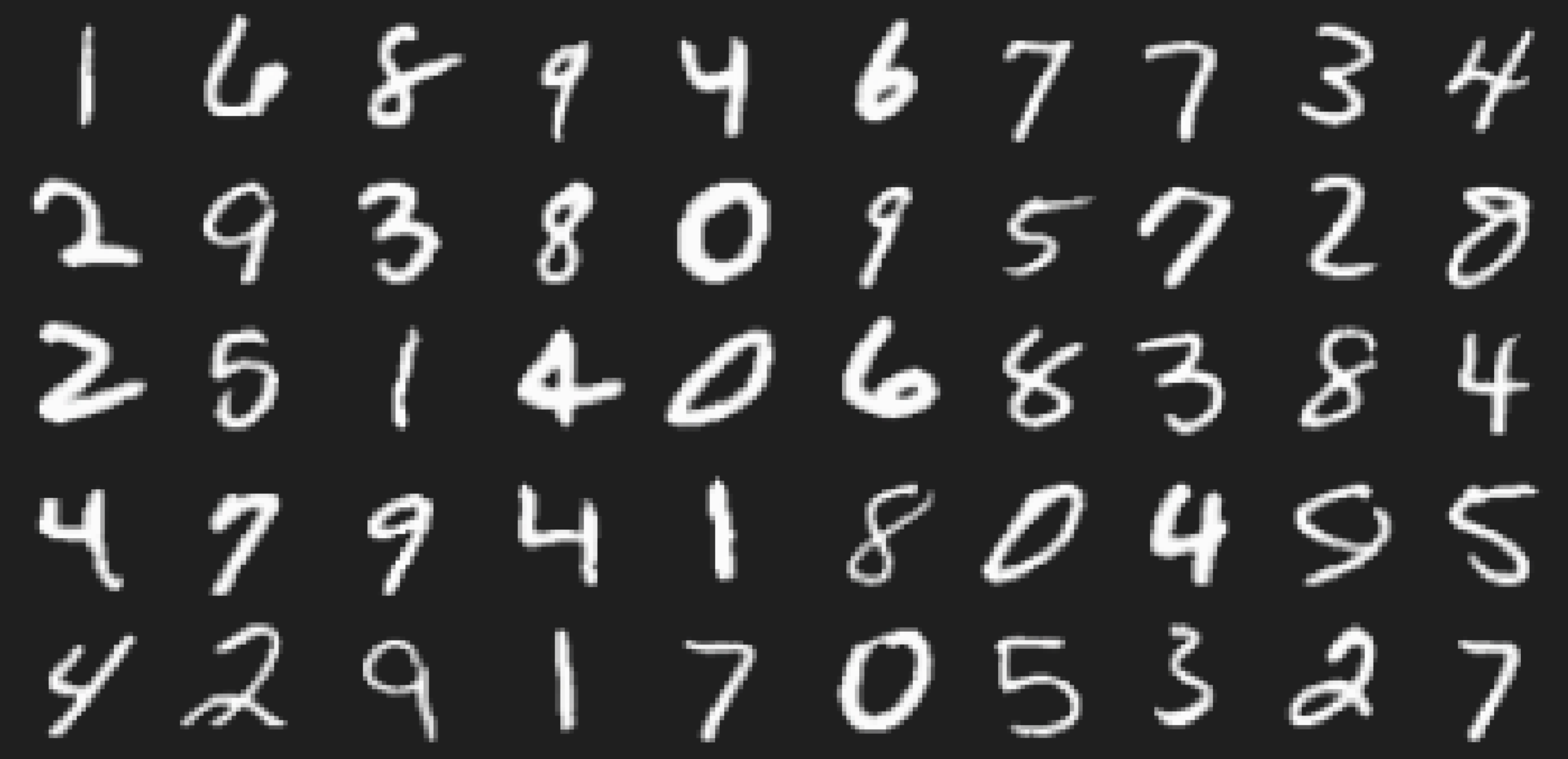}
\caption{MNIST samples obtained with the time-reversed scaled Ehrenfest process which was trained with $\mathcal{L}_\mathrm{OU}$.}
\label{fig: mnist}
\end{figure}

For a basic image modeling task, we consider the MNIST dataset, which consists of gray scale pixels and was resized to $32 \times 32$ to match the required input size of a U-Net neural network architecture\footnote{Taken from the repository \url{https://github.com/w86763777/pytorch-ddpm}.}, such that $d=32\times32=1024$ and $S=255$.
As before, we train our time-reversed Ehrenfest model by using the variants of the loss introduced in \Cref{sec: loss functions}.
In \Cref{fig: mnist} we display generated samples from a model trained with $\mathcal{L}_\mathrm{OU}$. The models with the other losses look equally good, so we omit them. For further details, we refer to \Cref{app: MNIST}.

\subsection{Image modeling with CIFAR-10}

As a more challenging task, we consider the CIFAR-10 data set, with dimension $d = 3 \times 32 \times 32 = 3072$, each taking $256$ different values \cite{krizhevsky2009learning}.
In the experiments we again compare our three different losses, however, realize that $\mathcal{L}_\mathrm{Gau\ss}$ did not produce satisfying results and had convergence issues, which might follow from numerical issues due to the exponential term appearing in \eqref{eq: Gauss loss}. Further, we consider three different scenarios: we train a model from scratch, we take the U-Net model that was pretrained in the state-continuous setting, and we take the same model and further train it with our state-discrete training algorithm (recall \Cref{rem: Ehrenfest as discrete DDPM}, which describes how to link the Ehrenfest process to DDPM). 

We display the metrics in \Cref{tab: CIFAR metrics}. When using only transfer learning, the different losses indicate different ways of incorporating the pretrained model, see \Cref{app: Ehrenfest - score based SDE}.
We realize that both losses produce comparable results, with small advantages for $\mathcal{L}_\mathrm{OU}$. Even without having invested much time in finetuning hyperparameters and sampling strategies, we reach competitive performance with respect to the alternative methods LDR \cite{campbell2022continuous} and D3PM \cite{austin2021structured}. Remarkably, even the attempt with transfer learning returns good results, without having applied any further training. For further details, we refer to \Cref{app: CIFAR}, where we also display more samples in Figures \ref{fig:cifar 10 taylor big 1}-\ref{fig:cifar 10 score big 2}.

\begin{figure}[ht]
    \centering
    \begin{minipage}{0.23\textwidth}
        \centering
        \includegraphics[width=\linewidth]{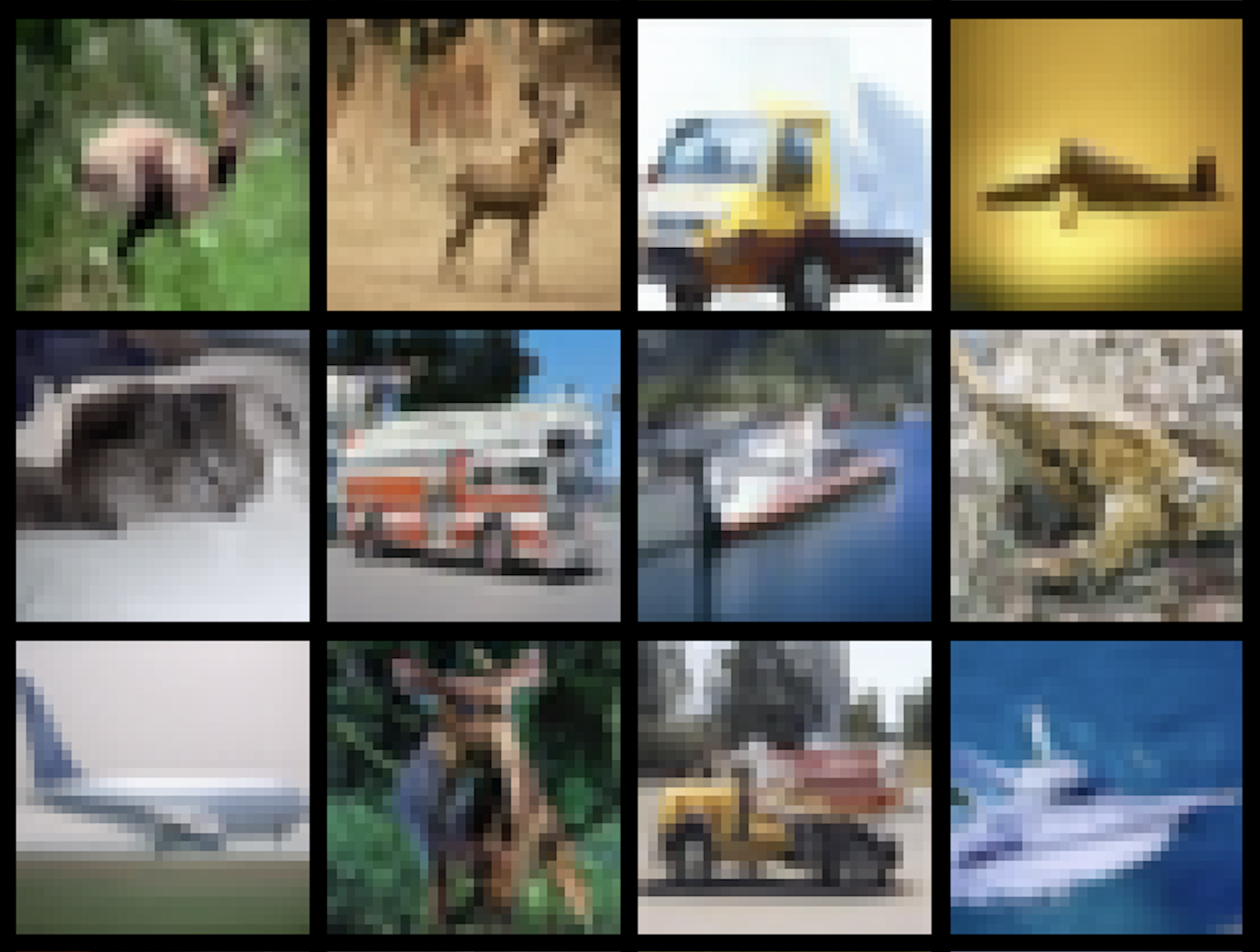} 
        \caption{CIFAR-10 samples from the Ehrenfest process with a pretrained model, further finetuned with $\mathcal{L}_\mathrm{OU}$.}
        \label{fig:figure1}
    \end{minipage}\hfill
    \begin{minipage}{0.23\textwidth}
        \centering
        \includegraphics[width=\linewidth]{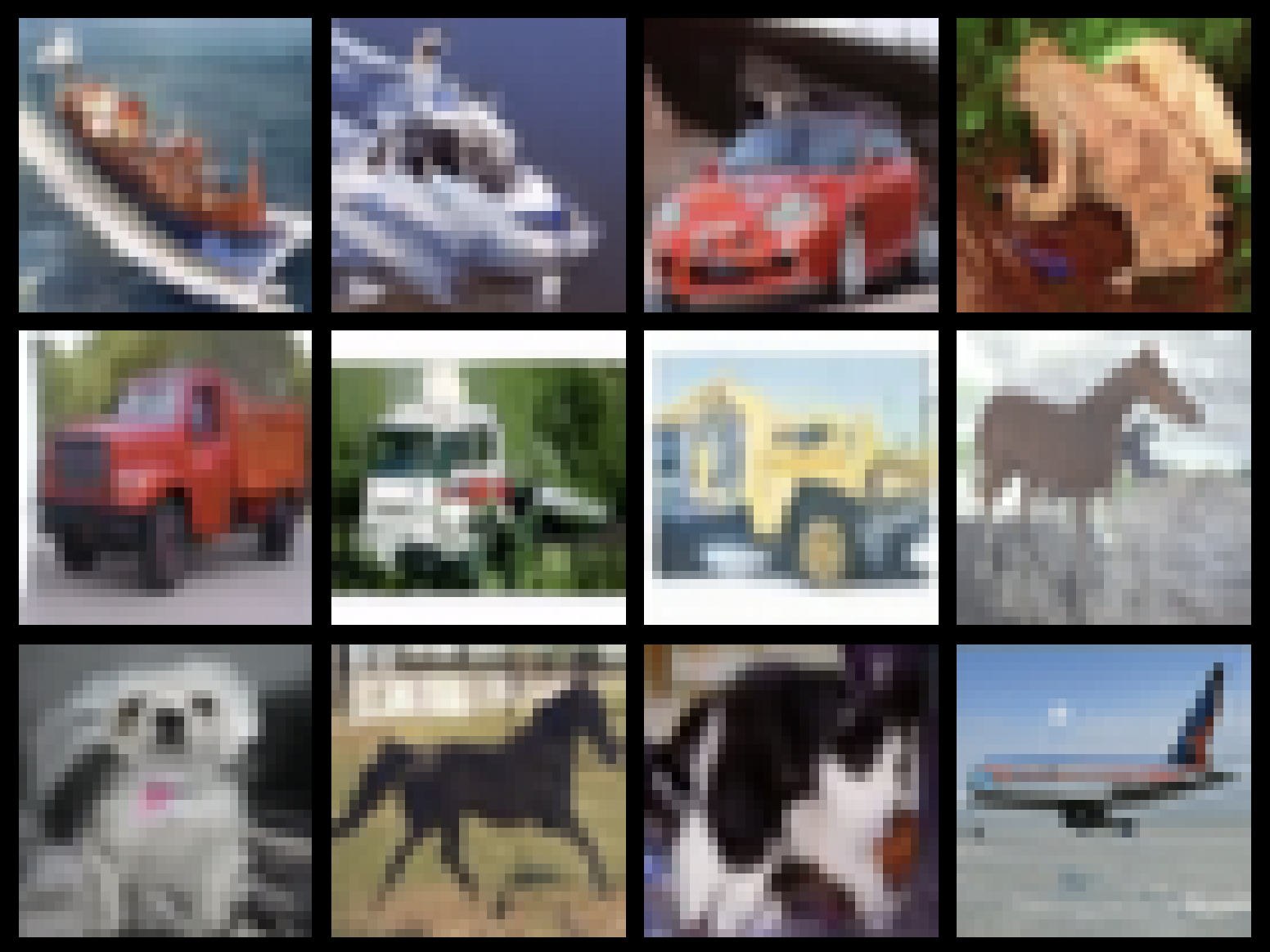} 
        \caption{CIFAR-10 samples from the Ehrenfest process with a pretrained model, further finetuned with $\mathcal{L}_\mathrm{Taylor}$.}
        \label{fig:figure2}
    \end{minipage}
\end{figure}

\begin{table}[H]
\centering
\begin{tabular}{clcc}
 & & IS ($\uparrow$) & FID ($\downarrow$) \\
\hline
Ehrenfest           & $\mathcal{L}_\mathrm{OU}$ & $8.75$ & $11.57$ \\ 
(transfer learning) & $\mathcal{L}_\mathrm{Taylor}$ & $8.68$ & $11.72$ \\
\hline
Ehrenfest           & $\mathcal{L}_\mathrm{OU}$ & $9.50$ & $5.08$\\ 
(from scratch)      & $\mathcal{L}_\mathrm{Taylor}$ & $9.66$ & $5.12$ \\
                    & $\mathcal{L}_\mathrm{Taylor2}$ & $9.40$ & $5.44$ \\
\hline
Ehrenfest    & $\mathcal{L}_\mathrm{OU}$ & $9.14$ & $6.63$ \\ 
(pretrained) & $\mathcal{L}_\mathrm{Taylor}$ & $9.06$ & $6.91$ \\
\hline
  & $\tau$-LDR (0) & $8.74$ & $8.10$ \\
Alternative & $\tau$-LDR (10) & $9.49$ & $3.74$ \\
methods & D3PM Gauss & $8.56$ & $7.34$ \\
                    & D3PM Absorbing & $6.78$ & $30.97$ \\
\hline
\end{tabular}
\caption{Performance in terms of Inception Score (IS) \cite{salimans2016improved} and Frechet Inception Distance (FID) \cite{heusel2017gans} on CIFAR-10 over $50.000$ samples. We compare two losses and consider three different scenarios: we train a model from scratch, we take the U-Net model that was pretrained in the state-continuous setting (called ``transfer learning'') or we take the same model and further train it with our state-discrete training algorithm (called ``pretraining'').}
\label{tab: CIFAR metrics}
\end{table}

\section{Conclusion}
\label{sec: conclusion}

In this work, we have related the time-reversal of discrete-space Markov jump processes to continuous-space score-based generative modeling, such that, for the first time, one can directly link models of the respective settings to one another. While we have focused on the theoretical connections, our numerical experiments demonstrate that we can already reach competitive performance with the new loss function that we proposed. We suspect that further tuning and the now possible transfer learning between discrete and continuous state space will further enhance the performance. On the theoretical side, we anticipate that the convergence of the time-reversed jump processes to the reversed SDE can be generalized even further, which we leave to future work.

\section*{Acknowledgements}

L.W. acknowledges support by the Federal Ministry of Education and Research (BMBF) for BIFOLD (01IS18037A). The research of L.R. has been partially funded by Deutsche Forschungsgemeinschaft (DFG) through the grant CRC 1114 ``Scaling Cascades in Complex Systems'' (project A05, project number 235221301). M.O. has been partially funded by Deutsche Forschungsgemeinschaft (DFG) through the grant CRC 1294 ``Data Assimilation'' (project number 318763901).

\section*{Impact statement}

The goal of this work is to advance the theoretical understanding of generative modeling based on stochastic processes, eventually leading to improvements in applications as well. While there are potential societal consequences of our work in principle, we do not see any concrete issues and thus believe that we do not specifically need to highlight any.

\bibliography{references}
\bibliographystyle{icml2024}

\newpage
\appendix
\onecolumn

\section{Proofs and additional statements}
\label{app: proofs}

In this section, we provide the proofs of the statements in the main text and state some additional lemmas, which are helpful for our arguments.

\begin{proof}[Proof of \Cref{lem: backward rates}]

First, we note that the derivation of the backward rates is known, see, e.g., \citet{campbell2022continuous}. For convenience, we repeat the essential part of the proof. Starting with the identity
\begin{equation}
    \cev{r}_t(y | x) p_t(x) = r_t(x | y) p_t(y),
\end{equation}
we can compute
\begin{subequations}
\begin{align}
    \cev{r}_t(y | x) &= \frac{p_t(y)}{p_t(x)} r_t(x | y) \\
    &=  \frac{\sum_{x_0 \in \Omega} p_{t|0}(y|x_0) p_0(x_0)}{p_{t}(x)}  r_t(x | y) \\
    &= \sum_{x_0\in \Omega} \frac{ p_{t|0}(y|x_0)}{ p_t(x | x_0)}  \frac{p_{t|0}(x | x_0) p_0(x_0)}{p_{t}(x)} r_t(x | y) \\
    &= \sum_{x_0\in \Omega} \frac{ p_{t|0}(y|x_0)}{ p_{t|0}(x | x_0)} \ p_{0|t}(x_0 | x) r_t(x | y)\\
    &= \mathbb{E}_{x_0 \sim p_{0|t}(x_0 | x)} \left[ \frac{ p_{t|0}(x|x_0)}{ p_{t|0}(y | x_0)} \right] r_t(x | y),
\end{align}
\end{subequations}
which shows the identity.
\end{proof}

\begin{lemma}[Score function as conditional expectation]
\label{lem: score as conditional expectation}

Consider the diffusion process $X_t$ defined by the SDE 
\begin{equation}
    \mathrm d X_t = b(X_t, s) \mathrm dt + \sigma(t) \mathrm d W_t, \qquad X_0 \sim p_\mathrm{data},
\end{equation}
with suitable drift function $b:\R^d \times [0, T] \to \R^d$ and diffusion coefficient $\sigma : [0,T] \to \R^{d\times d}$, let $p_t^\mathrm{SDE}$ be its marginal density and let $p^\mathrm{SDE}_{t|s}(x|y) := \mathbb{P}(X_t = x | X_s = y)$ for $t > s \ge 0$ be a transition probability. It then holds
    \begin{equation}
    \nabla_x \log p_t^\mathrm{SDE}(x) = \E_{x_0 \sim p^\mathrm{SDE}_{0|t}(x_0 | x)}\left[\nabla_x \log p^\mathrm{SDE}_{t | 0}(x |x_0) \right].
\end{equation}
\end{lemma}
\begin{proof}
    Noting the identity $p^\mathrm{SDE}_t(x) = \int_{\R^d} p^\mathrm{SDE}_{t | 0}(x|x_0) p_\mathrm{data}(x_0)\mathrm d x_0$, we can compute
    \begin{subequations}
    \begin{align}
        \nabla \log p^\mathrm{SDE}_t(x) &= \frac{\nabla_x p^\mathrm{SDE}_t(x)}{p^\mathrm{SDE}_t(x)} \\
        &= \frac{\int_{\R^d} \nabla_x \log p^\mathrm{SDE}_{t | 0}(x|x_0) p^\mathrm{SDE}_{t | 0}(x|x_0) p_\mathrm{data}(x_0)\mathrm d x_0}{\int_{\R^d} p^\mathrm{SDE}_{t | 0}(x|x_0) p_\mathrm{data}(x_0)\mathrm d x_0} \\
        &= \E_{x_0 \sim p^\mathrm{SDE}_{0|t}(x_0 | x)}\left[\nabla_x \log p^\mathrm{SDE}_{t | 0}(x |x_0) \right],
    \end{align}
    \end{subequations}
    where it holds $p^\mathrm{SDE}_{0|t}(x_0 | x) = \frac{p^\mathrm{SDE}_{t | 0}(x|x_0) p_\mathrm{data}(x_0)}{\int_{\R^d} p^\mathrm{SDE}_{t | 0}(x|x_0) p_\mathrm{data}(x_0)\mathrm d x_0}$ by Bayes' formula.
\end{proof}

\begin{lemma}[Conditional expectation as $L^2$ projection]
\label{lem: minimizer L2 error}
Let $A \in \R^{d}$ and $B \in \R$ be two random variables and let $\varphi \in C(\R^{d}, \R)$. Then the solution to 
\begin{equation}
\label{eq: minimizer L2 projection}
\varphi^* = \argmin_{\varphi \in C(\R^{d}, \R)} \E\left[\left(\varphi(A) - B \right)^2 \right]
\end{equation}
is given by
\begin{equation}
    \varphi^*(a) = \E[B | A=a].
\end{equation}
\end{lemma}
\begin{proof}

Let $\varphi^C(a) = \E[B | A=a]$. We compute
\begin{subequations}
\begin{align}
 \mathbb{E}\left[(\varphi(A)-B)^2\right] &= \mathbb{E}\left[(\varphi(A) - \varphi^C(A) + \varphi^C(A) -B)^2\right] \\
&= \mathbb{E}\left[(\varphi(A) - \varphi^C(A))^2\right] + \mathbb{E}\left[(\varphi^C(A) - B)^2\right] - 2\, \mathbb{E}\left[(\varphi(A) - \varphi^C(A))(\varphi^C(A) - B)\right],
\end{align}
\end{subequations}
which is minimized by $\varphi = \varphi^C$ since the last term is equal to\footnote{Here the notation $\E_A$ refers to the expectation over $A$, whereas $\E_{B|A}$ refers to the expectation over $B$ conditional on $A$.}
\begin{align}
 \mathbb{E}_A\left[ \mathbb{E}_{B|A}\left[(\varphi(A) - \varphi^C(A))(\varphi^C(A) - B)\right] \right] =  \mathbb{E}_A\left[ (\varphi(A) - \varphi^C(A)) \mathbb{E}_{B|A}\left[(\varphi^C(A) - B)\right] \right] = 0.
\end{align}
Therefore $\varphi^* = \varphi^C$.
\end{proof}

\begin{proof}[Proof of \Cref{lem: forward transition prob Ehrenfest}]
We consider the Ehrenfest process as defined in \eqref{eq: definition Ehrenfest}, assuming that it starts at $E_S(0) = x_0$. We can write the process as
\begin{equation}
    E_S(t) = \sum_{i=1}^S Z_i(t) = \sum_{Z_i(0) = 0}Z_i(t) + \sum_{Z_i(0) = 1}Z_i(t) =: E_{0,S}(t) + E_{1,S}(t),
\end{equation}  
where $E_{1, S}$ is a sum of $x_0$ independent Bernoulli random variables $Z_i$ with 
\begin{equation}
    f(t) := \mathbb{P}(Z_i(t) = 1 | Z_i(0)=1) = \frac{1}{2}\left(1 + e^{-t}\right),
\end{equation}
and where $E_{0, S}$ is the sum of $S-x_0$ random variables $Z_i$ with $\mathbb{P}(Z_i(t) = 1|Z_i(0)=0) = 1-f(t) = \frac{1}{2}\left(1 - e^{-t}\right)$. Thus, both $E_{0, S}$ and $E_{1, S}$ are binomial random variables distributed as
 \begin{align}
E_{0, S}(t) \sim B(S - x_0, 1 - f(t)), \qquad E_{1, S}(t) \sim B(x_0, f(t)).
 \end{align}
\end{proof}

\begin{proof}[Proof of \Cref{prop: jump moment reversed Ehrenfest}]
    We first recall the scaled Ehrenfest process from \eqref{eq: scaled Ehrenfest},
    \begin{equation}
        \widetilde{E}_S(t) := \frac{2}{\sqrt{S}}\left(E_{S}(t) -  \frac{S}{2}\right),
    \end{equation}
    and note that $\widetilde{E}_S \in \left\{-\sqrt{S},-\sqrt{S}+\frac{2}{\sqrt{S}},\dots,\sqrt{S} \right\}$, where the birth-death transitions transform from $\pm 1$ in $E_S$ to $\pm \frac{2}{\sqrt{S}}$ in its scaled version $\widetilde{E}_S$. Accordingly, the reverse rates from \Cref{lem: backward rates} translate to
    \begin{equation}
        \cev{r}_t\left(x \pm \frac{2}{\sqrt{S}} \bigg| x\right) = \E_{x_0 \sim p_{0|t}(x_0 | x)}\left[ \frac{p_{t|0}\left(x \pm \frac{2}{\sqrt{S}} \Big| x_0\right)}{p_{t|0}(x | x_0)} \right] r\left(x \bigg| x \pm \frac{2}{\sqrt{S}}\right).
    \end{equation}
    Let us introduce the notation (which slightly deviates from the notation in \Cref{prop: jump moment reversed Ehrenfest})
    \begin{equation}
        \Delta_\delta p_{t|0}(x | x_0) := \frac{ p_{t|0}(x + \delta | x_0) -  p_{t|0}(x  | x_0)}{\delta}
    \end{equation} 
    and note the identity
    \begin{equation}
        p_{t|0}(x + \delta | x_0) = p(x | x_0) + \delta \, \Delta_\delta p_{t|0}(x | x_0),
    \end{equation}
    which is sometimes called \textit{Newton's series for equidistant nodes} and can be seen as a discrete analog of a Taylor series, where, however, terms of order higher than one vanish. We can now compute the first jump moment
\begin{subequations}
\begin{align}
    \cev{b}(x) &:= \sum_{\delta \in \left\{-\frac{2}{\sqrt{S}}, \frac{2}{\sqrt{S}}\right\}} \delta \, \cev{r}(x + \delta | x) =  \sum_{\delta \in \left\{-\frac{2}{\sqrt{S}}, \frac{2}{\sqrt{S}}\right\}} \delta \, \E_{x_0 \sim p_{0|t}(x_0 | x)}\left[ \frac{p_{t|0}(x + \delta | x_0)}{p_{t|0}(x | x_0)} \right] r(x | x + \delta) \\
    &= \sum_{\delta \in \left\{-\frac{2}{\sqrt{S}}, \frac{2}{\sqrt{S}}\right\}} \delta \, \E_{x_0 \sim p_{0|t}(x_0 | x)}\left[ 1 + \frac{\delta \, \Delta_\delta p_{t|0}(x | x_0)}{p_{t|0}(x | x_0)} \right] r(x | x + \delta) \\
    &= \sum_{\delta \in \left\{-\frac{2}{\sqrt{S}}, \frac{2}{\sqrt{S}}\right\}} \delta \,   r(x | x + \delta) + \sum_{\delta \in \left\{-\frac{2}{\sqrt{S}}, \frac{2}{\sqrt{S}}\right\}} \delta^2 \,  \E_{x_0 \sim p_{0|t}(x_0 | x)}\left[\frac{\Delta_\delta p_{t|0}(x | x_0)}{p_{t|0}(x | x_0)} \right] r(x | x + \delta) \\
    &= \sum_{\delta \in \left\{-\frac{2}{\sqrt{S}}, \frac{2}{\sqrt{S}}\right\}} \delta \,   r(x | x + \delta) + \E_{x_0 \sim p_{0|t}(x_0 | x)}\left[\frac{\Delta_{\bar{\delta}} p_{t|0}(x | x_0)}{p_{t|0}(x | x_0)} \right] \sum_{\delta \in \left\{-\frac{2}{\sqrt{S}}, \frac{2}{\sqrt{S}}\right\}} \delta^2 \,   r(x | x + \delta) + o(\bar{\delta})\\
    &= -b(x)+ D(x)\E_{x_0 \sim p_{0|t}(x_0 | x)}\left[\frac{\Delta_{\bar{\delta}} p_{t|0}(x | x_0)}{p_{t|0}(x | x_0)} \right]  + o(\bar{\delta}),
\end{align}
\end{subequations}
where for $\bar{\delta}:=\frac{2}{\sqrt{S}}$ we have used that $\Delta_{\bar{\delta}}p_{0|t}(x|x_0) = \Delta_{-\bar{\delta}}p_{0|t}(x|x_0) + o(\bar{\delta})$ since
\begin{subequations}
\begin{align}
     \frac{p(x | x_0) - p(x - \bar{\delta} | x_0)}{\bar{\delta}} &= \frac{p(x + \bar{\delta}| x_0) - p(x | x_0)}{\bar{\delta}} +\frac{2 \, p(x | x_0) - p(x + \bar{\delta}| x_0) - p(x - \bar{\delta} | x_0)}{\bar{\delta}}\\
    &=\frac{p(x + \bar{\delta}| x_0) - p(x | x_0)}{\bar{\delta}} + o(\bar{\delta}),
\end{align}
\end{subequations}

as well as
\begin{equation}
    \sum_{\delta \in \left\{-\frac{2}{\sqrt{S}}, \frac{2}{\sqrt{S}}\right\}} \delta \, r(x | x + \delta) = -\sum_{\delta \in \left\{-\frac{2}{\sqrt{S}}, \frac{2}{\sqrt{S}}\right\}} \delta \, r(x+ \delta | x ) + o(S^{-1/2}),
\end{equation}
and 
\begin{equation}
\label{eq: rate reversal}
    \sum_{\delta \in \left\{-\frac{2}{\sqrt{S}}, \frac{2}{\sqrt{S}}\right\}} \delta^2 r(x | x + \delta) = \sum_{\delta \in \left\{-\frac{2}{\sqrt{S}}, \frac{2}{\sqrt{S}}\right\}} \delta^2 r(x+ \delta | x ) + o(S^{-1}),
\end{equation}
since
\begin{subequations}
\begin{align}
    r\left(x \bigg| x \pm \frac{2}{\sqrt{S}}\right) &= \frac{\sqrt{S}}{4}\left(\sqrt{S} \pm \left(x \pm \frac{2}{\sqrt{S}}\right) \right) =  \frac{\sqrt{S}}{4}\left(\sqrt{S} \pm x  \right) \pm \frac{1}{2} \\
    &= r\left(x \mp \frac{2}{\sqrt{S}}  \bigg| x\right) \pm \frac{1}{2}.
\end{align}
\end{subequations}

\end{proof}

\section{Background on time-continuous Markov jump processes}

In this section we will provide some background on continuous-time, discrete-space  Markov jump processes. 

\subsection{A brief introduction to Markov jump processes}
\label{sec: intro Markov jump processes}

In this section we will give a brief introduction to time-continuous Markov processes on a discrete state space, which is based on a summary in \citet[Section 2.2]{metzner2008transition}. We refer the interested reader to \citet{gardiner1985handbook,van1992stochastic,bremaud2013markov} for further details.

We denote with $X_t$ an $\Omega$-valued stochastic process on a discrete (countable) state space $\Omega$ with a continuous time parameter $0 \le t < \infty$. The process is called a Markov process if for all times $t_{k+1} > t_k \ge \dots \ge t_0 = 0$ and for any $x_{k+1}, \dots, x_0 \in \Omega$ it holds
\begin{equation}
    \P\left(X_{t_{k+1}} = x_{k+1} \big| X_{t_{k}} = x_{k}, \cdots, X_{t_{0}} = x_{0}\right) = \P\left(X_{t_{k+1}} = x_{k+1} \big| X_{t_{k}} = x_{k}\right).
\end{equation}
The process is called \textit{homogeneous} if the transition probability only depends on the time increment $t_{k+1} - t_{k}$. We denote with
\begin{equation}
    p_{t | s}(x | y) = \P(X_t = x | X_s = y)
\end{equation}
the transition probability for times $t > s > 0$ and define the matrix
\begin{equation}
    P(t) := \left(p_{t | 0}(x | y)\right)_{x,y\in\Omega}.
\end{equation}
$P(t)$ is a stochastic matrix, i.e.
\begin{equation}
    p_{t | 0}(x | y) \ge 0, \qquad \sum_{y \in \Omega} p_{t | 0}(x | y) = 1,
\end{equation}
 for each time $t \ge 0$ and each $x, y \in \Omega$. The family of transition matrices $\left\{P(t)\right\}_{t \ge 0}$ is called transition semi-group since it obeys the \textit{Chapman-Kolmogorov equation}
 \begin{equation}
     P(t + s) = P(t)P(s)
 \end{equation}
for $s, t \ge 0$ with $P(0) = \operatorname{Id}$.

A local characterization of the transition semigroup of a Markov jump process can be obtained by considering the infinitesimal changes of the transition probabilities. One can show that the limit
\begin{equation}
    R = \lim_{t \to 0^+}\frac{P(t) - \operatorname{Id}}{t}
\end{equation}
exists (entrywise), which is sometimes written as 
\begin{equation}
    p_{t+\Delta t| t}(x | y) = \delta_{x, y} + r_t(x| y) \Delta t + o(\Delta t),
\end{equation}
cf. equation \eqref{eq: transition Markov jump} in \Cref{sec: time-reversed jump processes}.
The matrix $R = \left(r(x | y)\right)_{x, y \in \Omega}$ is called infinitesimal generator of the transition semigroup $\left\{P(t)\right\}_{t \ge 0}$ because it ``generates'' the transition semigroup via the relation
\begin{equation}
    P(t) = e^{t R} = \sum_{n=0}^\infty \frac{t^n}{n!} R^n. 
\end{equation}
One can show that
\begin{equation}
    0 \le r(x | y) < \infty, \qquad \sum_{y \in \Omega}r(x | y) = 0,
\end{equation}
for all $x, y \in \Omega$ with $x \neq y$, and we can interpret $r(x | y)$ as a transition rate from state $y$ to $x$, measuring the average number of transitions per unit time. The diagonal elements of $R$ are defined as
\begin{equation}
    r(x | x) = - \sum_{y \neq x} r(x | y)
\end{equation}
for each $x \in \Omega$. Analog to the state-continuous case, it holds the \textit{backward Kolmogorov equation} for the conditional expectation $\psi(t) = \left(\E\left[ f(X_t) | X_0 = x \right]\right)^\top_{x\in\Omega}$ for an observable $f:\Omega \to \R$, namely
\begin{equation}
    \frac{\mathrm d}{\mathrm d t} \psi(t) = R \psi(t), \qquad \psi(0) = (f(x))^\top_{x\in\Omega}.
\end{equation}
Further, for the vector of state probabilities $p(t):= \left(p_t(x)\right)_{x \in \Omega}$, recalling the notation $p_t(x) := \P(X_t = x) $, it holds the \textit{forward Kolmogorov equation}, also known as \textit{Master equation}, namely
\begin{equation}
    \frac{\mathrm d}{\mathrm d t} p(t) = p(t) R.
\end{equation}
For the transition densities we have
\begin{equation}
    \frac{\mathrm d}{\mathrm d t} p_{t|0}(x | y) = \sum_{z\in \Omega}p_{t|0}(z|y)R(z, x),
\end{equation}
or in matrix notation
\begin{equation}
\label{eq: master equation}
    \frac{\mathrm d}{\mathrm d t} P(t) = P^\top(t) R.
\end{equation}

It can be solved as
\begin{equation}
\label{eq: solution of master equation}
    \left(p_{t|0}(x | y)\right)_{x, y \in \Omega} = \exp\left(\int_0^t \left(r_s(x | y)\right)_{x, y \in \Omega} \mathrm ds \right) ,
\end{equation}
where $\exp$ is the matrix exponential.

\subsection{Time-transformation of Markov jump processes}
\label{app: time transformation}

Note that we can always transform a Markov jump process with a time dependent rate $r_t$ into one with a time independent rate $r$ using a time transformation. This can be seen by looking at the master equation defined in \eqref{eq: master equation}, namely
\begin{equation}
    \frac{\mathrm d}{\mathrm d t} P(t) = P^\top(t) R_t,
\end{equation}
where now the rate matrix $R_t$ is time-dependent. For simplicity, let us assume $R_t = \lambda_t R$, where $\lambda : [0, T] \to \R$ and $R$ is time-independent. We can now introduce the new time $\tau = \tau(t)$ and compute
\begin{align}
    \frac{\mathrm d}{\mathrm d t} P(\tau(t)) =\frac{\mathrm d P(\tau)}{\mathrm d \tau} \frac{\mathrm d \tau}{\mathrm d t} = P^\top(\tau(t)) \lambda_t R.
\end{align}
Now, choosing $\frac{\mathrm d \tau}{\mathrm d t} = \lambda_t$ and thus $\tau(t) = \int_0^t \lambda_s \mathrm ds$ (where we have assumed $\tau(0) = 0$), yields the equation
\begin{align}
    \frac{\mathrm d }{\mathrm d \tau}P(\tau)  = P^\top(\tau) R,
\end{align}
where now the rate matrix does not depend on time anymore.

\subsection{Convergence of Markov jump processes}
\label{app: convergence of jump processes}

The convergence of Markov jump processes to SDEs in the limit of large state spaces (with appropriately scaled jump sizes) has formally been studied via the Kramers-Moyal expansion \cite{gardiner1985handbook,van1992stochastic}. For more rigorous results, we refer, e.g., to \cite{kurtz1972relationship,kurtz1981approximation}.

One can get some intuition by looking at the first two jump moments of the Markov jump process. The first jump moment is defined as
\begin{subequations}
\begin{align}
    b(x) &:= \lim_{\Delta t \to 0} \frac{1}{\Delta t}\E\left[M(t + \Delta t) - M(t) | M(t) = x \right] \\
    &= \lim_{\Delta t \to 0}\frac{1}{\Delta t}\left(\sum_{y \in \Omega} y \, p_{t + \Delta t|t}(y | x) - x \sum_{y \in \Omega} p_{t + \Delta t|t}(y | x) \right) \\
    &= \lim_{\Delta t \to 0}\frac{1}{\Delta t}\left(\sum_{y \in \Omega; y\ne x} (y-x) p_{t + \Delta t|t}(y | x) \right) \\
    &= \sum_{y \in \Omega; y\ne x} (y-x) r(y | x).
\end{align}
\end{subequations}
Similarly, the second jump moment is defined as
\begin{subequations}
\begin{align}
    D(x) &:=  \lim_{\Delta t \to 0} \frac{1}{\Delta t}\E\left[\left(M(t + \Delta t) - M(t)\right)\left(M(t + \Delta t) - M(t)\right)^\top | M(t) = x \right] \\
    &=  \sum_{y \in \Omega; y\ne x} (y-x)(y-x)^\top r(y | x).
\end{align}
\end{subequations}
Note that the drift and diffusion coefficient for SDEs are defined analogously. 

\section{Computational aspects}

In this section we comment on computational aspects of the time-reversal of Markov jump processes.

\subsection{Approximation of the conditional expectation}
\label{sec: computational aspects cond exp}

For the approximation of the conditional expectation appearing in the backward rates from \Cref{lem: backward rates} we propose \Cref{alg: conditional expectation} and for sampling from a time-reversed process with approximated backward rates we propose \Cref{alg: sampling}. 

\begin{algorithm}[t!]
\begin{algorithmic}
\INPUT Batch size $K$, gradient steps $M$, two neural networks $\varphi_b$ and $\varphi_d$ with initial parameters $\theta^{(0)}_b$ and $\theta^{(0)}_d$, respectively, approximations $\widetilde{p}_{0|t} \approx p_{0|t}$ (typically either by \eqref{eq: convolution of binomials} or \eqref{eq: approx of forward transition by Gaussian}).
\OUTPUT Approximations of the conditional expectations appearing in \eqref{eq: backward rates}.
\FOR{$m\gets 0,\dots, M-1$}
\STATE Sample data points $x_0^{(1)}, \dots, x_0^{(K)} \sim p_\mathrm{data}$.
\STATE Sample terminal times $t_1, \dots, t_K \sim \mathcal{U}(0, T)$.
\STATE Simulate $x^{(k)}_{t_k}$ for each $k \in \{1, \dots, K \}$ according to the forward (scaled) Ehrenfest process \eqref{eq: scaled Ehrenfest}. Note that every dimension can be sampled independently and simulation-free as binomial random variables, see \Cref{lem: forward transition prob Ehrenfest}.
\STATE Compute two losses:
\begin{subequations}
\begin{align}
\widehat{\mathcal{L}}_b(\theta_b^{(m)}) &= \frac{1}{K d} \sum_{k=1}^K\sum_{i=1}^d\left(\varphi_b^{(i)}(x^{(k)}_{t_k}, t_k) - \frac{ \widetilde{p}^{(i)}_{t_k|0}\left(x^{(i),(k)}_{t_k} + \frac{2}{\sqrt{S}}\big|x_0^{(i),(k)}\right)}{ \widetilde{p}^{(i)}_{t_k|0}\left(x^{(i),(k)}_{t_k} \big| x_0^{(i),(k)}\right)} \right)^2\\
\widehat{\mathcal{L}}_d(\theta_d^{(m)}) &= \frac{1}{K d} \sum_{k=1}^K\sum_{i=1}^d\left(\varphi_d^{(i)}(x^{(k)}_{t_k}, t_k) - \frac{ \widetilde{p}^{(i)}_{t_k|0}\left(x^{(i),(k)}_{t_k} - \frac{2}{\sqrt{S}}\big|x_0^{(i),(k)}\right)}{ \widetilde{p}^{(i)}_{t_k|0}\left(x^{(i),(k)}_{t_k} \big| x_0^{(i),(k)}\right)} \right)^2
\end{align}
\end{subequations}
\STATE Do gradient descent:
\begin{subequations}
\begin{align}
\theta_b^{(m + 1)} &\gets \operatorname{step}\left( \theta^{(m)}, \nabla \widehat{\mathcal{L}}_b(\theta_b^{(m)}) \right) \\
\theta_d^{(m + 1)} &\gets \operatorname{step}\left( \theta^{(m)}, \nabla \widehat{\mathcal{L}}_d(\theta_d^{(m)}) \right)
\end{align}
\end{subequations}
\ENDFOR
\end{algorithmic}
\caption{Approximation of conditional expectation for the Ehrenfest process.}
\label{alg: conditional expectation}
\end{algorithm}

\begin{algorithm}[t!]
\begin{algorithmic}
\INPUT Rate of the forward process $r_t$, approximation of the conditional expectations in \eqref{eq: backward rates} via $\varphi_b$ and $\varphi_b$ for the birth and death transitions, respectively, approximation of terminal distribution $p_\mathrm{ref} \approx p_T$, leaping time $\tau > 0$.
\OUTPUT Data sample that is approximately distributed according to $p_\mathrm{data}$.
\STATE Sample $x_T  \sim p_\mathrm{ref}$.
\STATE Set $t \gets T$.
\WHILE{$t > 0$}
\FOR{$i = 1, \dots, d$}
\STATE Compute backward rates:
  \begin{align}
        \cev{r}_t^{(i)}\left(x^{(i)} + \frac{2}{\sqrt{S}} \bigg| x^{(i)}\right) &= \varphi_b^{(i)}(x, t) \, r^{(i)}_t\left(x^{(i)} \bigg| x^{(i)} + \frac{2}{\sqrt{S}}\right)\\ 
        \cev{r}_t^{(i)}\left(x^{(i)} - \frac{2}{\sqrt{S}} \bigg| x^{(i)}\right) &= \varphi_d^{(i)}(x, t) \,  r^{(i)}_t\left(x^{(i)} \bigg| x^{(i)} - \frac{2}{\sqrt{S}}\right)
  \end{align}
  \STATE Draw Poission random variable:
  \begin{align}
         \rho_{b}^{(i)} &\sim \operatorname{Pois}\left(\tau \, \cev{r}_t^{(i)}\left(x^{(i)} + \frac{2}{\sqrt{S}} \Big| x^{(i)}\right) \right) \\
 \rho_{d}^{(i)} &\sim \operatorname{Pois}\left(\tau \, \cev{r}_t^{(i)}\left(x^{(i)} - \frac{2}{\sqrt{S}} \Big| x^{(i)}\right) \right)
  \end{align}

  \STATE Do leaping step:
    \begin{align}
   x_{t-\tau}^{(i)} &\gets x_{t}^{(i)} + \rho_{b}^{(i)} \frac{2}{\sqrt{S}} - \rho_{d}^{(i)} \frac{2}{\sqrt{S}} \\
   x_{t-\tau}^{(i)} &\gets \operatorname{Clamp}(x_{t-\tau}^{(i)}, -\sqrt{S}, \sqrt{S})
     \end{align}
\ENDFOR
\STATE $t \gets t - \tau$
\ENDWHILE
\STATE Return $x_0$.
\end{algorithmic}
\caption{Sampling from data distribution $p_\mathrm{data}$.}
\label{alg: sampling}
\end{algorithm}

\subsection{Learning the reversed transition probability}
\label{sec: Learning the reversed transition probability}

An alternative way to approximate the backward rates specified in \Cref{lem: backward rates} is to approximate the reversed transition probability $p_{0|t}(x_0|x)$ with a tractable distribution $p_{0|t}^{\theta}(x_0 | x)$. To this end, one can consider the loss
\begin{equation}
\label{def: ML loss}
    \mathcal{L}(\theta) := -\E_{t\sim\mathcal{U}(0, T), x_0 \sim p_\mathrm{data}, x \sim p_{t|0}(x|x_0)}\left[\log  p_{0|t}^\theta(x_0 | x) \right].
\end{equation}
The following lemma motivates this loss (cf. Proposition 8 in \citet{campbell2022continuous}).

\begin{lemma}
    It holds
    \begin{equation}
        \mathcal{L}(\theta) = \E_{t\sim\mathcal{U}(0, T), x \sim p_{t}(x)}\left[D_\mathrm{KL}\left(p_{0|t}(x_0 | x)|p_{0|t}^{\theta}(x_0 | x) \right) \right] + C,
    \end{equation}
    where $C$ is a constant that does not depend on $\theta$.
\end{lemma}

\begin{proof}
    Let $C$ be a constant that does not depend on $\theta$. We can compute
    \begin{subequations}
    \begin{align}
         \E_{t\sim\mathcal{U}(0, T), x \sim p_{t}(x)}&\left[D_\mathrm{KL}\left(p_{0|t}(x_0 | x)\Big|p_{0|t}^{\theta}(x_0 | x) \right) \right] \\
         &=  \E_{t\sim\mathcal{U}(0, T), x \sim p_{t}(x)}\left[ \E_{x_0\sim p_{0|t}(x_0|x)} \left[ \log \frac{p_{0|t}(x_0 | x)}{p_{0|t}^{\theta}(x_0 | x)} \right] \right] \\
         &= \E_{t\sim\mathcal{U}(0, T), x \sim p_{t}(x),x_0\sim p_{0|t}(x_0|x)} \left[ \log \frac{p_{0|t}(x_0 | x)}{p_{0|t}^{\theta}(x_0 | x)} \right]  \\
         &= -\E_{t\sim\mathcal{U}(0, T), x_0 \sim p_\mathrm{data}, x \sim p_{t|0}(x|x_0)} \left[ \log p_{0|t}^{\theta}(x_0 | x) \right] - C,
    \end{align}
    \end{subequations}
    where we used the tower property of conditional expectations and the identity $p_t(x)p_{0|t}(x_0|x)=p_\mathrm{data}(x_0)p_{t|0}(x|x_0)$. Noting the definition of $\mathcal{L}$ in \eqref{def: ML loss} concludes the proof.
\end{proof}

The above guarantees that $p_{0|t}^{\theta}(x_0 | x) = p_{0|t}(x_0 | x)$ if and only if $\mathcal{L}(\theta) = 0$.

We therefore can use \Cref{alg: reverse transition probabilities} for approximating the backward rates and \Cref{alg: sampling with backward transition densities} for sampling the time-reversed process.

Note that all probabilities are probabilities on the discrete set $\Omega \subset \mathbb{Z}$, fulfilling e.g. $ \sum_{x_0 \in \Omega} p_{0|t}(x_0 | x) = 1$ for all $x \in \Omega$ and $t \in [0,T]$. Specifically, $(p_{0|t}(x_0 | x))_{x_0, x \in \Omega} \in [0, 1]^{(S+1) \times (S+1)}$ is a (stochastic) matrix for all $t \in [0, T]$. In practice, however, we often model $p_{0|t}(x_0 | x)$ with a continuous distribution, parametrized by a neural network, e.g. 
\begin{equation}
    \widetilde{p}^\theta_{0|t}(x_0 | x) := \mathcal{N}(x_0; \mu^\theta(x, t), \Sigma^\theta(x, t)),
\end{equation}
where mean and covariance are learned with a neural network $\varphi : \mathbb{R} \times [0, T] \to \mathbb{R}^2$, i.e.
\begin{equation}
    \varphi^\theta(x, t) = (\mu^\theta(x, t), \Sigma^\theta(x, t))^\top.
\end{equation}
In order to recover probabilities on a discrete set, we use the cumulative distribution function 
\begin{equation}
    \Phi(z; x, t) := \int_{-\infty}^z \widetilde{p}^\theta_{0|t}(x_0 | x) \mathrm d x_0,
\end{equation}
which is analytically available for a Gaussian. For a discrete state $x_0 \in \Omega$ we then approximate the probability via
\begin{equation}
    p_{0|t}(x_0 | x) \approx p^\theta_{0|t}(x_0 | x) :=  \Phi(x_0; x, t) - \Phi(x_0 - 1; x, t)
\end{equation}
for $x_0 \in \Omega \setminus \{ 0, S \}$. For $x_0 = 0$ we consider $p^\theta_{0|t}(x_0 | x) :=  \Phi(x_0; x, t)$ and for $x_0 = S$ we consider $p^\theta_{0|t}(x_0 | x) :=  \Phi(-\infty; x, t) - \Phi(x_0 - 1; x, t)$.

For the forward probabilities $p_{t|0}$, we can either compute the matrix
\begin{equation}
    \left(p_{t|0}(x | x_0)\right)_{x_0, x \in \Omega} = \exp\left(\int_0^t \left(r_s(x | x_0)\right)_{x_0, x \in \Omega} \mathrm ds \right) ,
\end{equation}
where $\exp$ is the matrix exponential, or we can approximate $p_{t|0}(x | x_0)$ with Gaussians due to the convergence properties of the Ehrenfest process.

\begin{algorithm}[t!]
\begin{algorithmic}
\INPUT Batch size $K$, gradient steps $M$, parametrization $p_{0|t}^{\theta^{(0)}}$ with initial parameters $\theta^{(0)}$. 
\OUTPUT $p_{0|t}^{\theta^{(M)}} \approx p_{0|t}$.

\FOR{$m\gets 0,\dots, M-1$}
\STATE Sample data point $x_0^{(1)}, \dots, x_0^{(K)} \sim p_\mathrm{data}$.
\STATE Sample terminal times $t_1, \dots, t_K \sim \mathcal{U}(0, T)$.
\STATE Simulate $x^{(k)}_{t_k}$ for each $k \in \{1, \dots, K \}$ according to the forward process.
\STATE Compute $\widehat{\mathcal{L}}(\theta^{(m)}) = -\frac{1}{K}\sum_{k=1}^K \log p_{0|t}^{\theta^{(m)}}(x_0^{(k)} | x^{(k)}_{t_k})$
\STATE $\theta^{(m + 1)} \gets \operatorname{step}\left( \theta^{(m)}, \nabla \widehat{\mathcal{L}}(\theta^{(m)}) \right)$
\ENDFOR
\end{algorithmic}
\caption{Approximation of reverse transition probability}
\label{alg: reverse transition probabilities}
\end{algorithm}

\begin{algorithm}[t!]
\begin{algorithmic}
\INPUT Rate of the forward process $r_t$, approximation of forward transition probabilities $\widetilde{p}_{t|0} \approx p_{t|0}$, approximation of reverse transition probabilities $p_{0|t}^{\theta} \approx p_{0|t}$, approximation of terminal distribution $p_\mathrm{ref} \approx p_T$.
\OUTPUT Data sample that is approximately distributed according to $p_\mathrm{data}$.

\STATE Sample $x_T  \sim p_\mathrm{ref}$.
\STATE Simulate birth-death process from time $t = T$ to $t = 0$ with backward rates
  \begin{equation}
      \cev{r}_t(x \pm 1 | x) = \sum_{x_0 \in \Omega} p_{0|t}^{\theta}(x_0 | x) \frac{\widetilde{p}_{t|0}(x \pm 1 | x_0)}{\widetilde{p}_{t|0}(x | x_0)} r_t(x | x \pm 1).
  \end{equation}
\STATE Return $x_0$.
\end{algorithmic}
\caption{Sampling from data distribution $p_\mathrm{data}$.}
\label{alg: sampling with backward transition densities}
\end{algorithm}

\section{Numerical details}

In this section we elaborate on numerical details regarding our experiments in \Cref{sec: experiments}.

\subsection{A tractable Gaussian toy example}
\label{app: Gaussian mixture score}

In order to illustrate the properties of the Ehrenfest process, we consider the following toy example. Let us start with the SDE setting and consider the data distribution
\begin{equation}
    p_\mathrm{data}(x) = \sum_{m=1}^M \gamma_m \mathcal{N}(x; \mu_m, \Sigma_m),
\end{equation}
where $\sum_{m=1}^M \gamma_m = 1$ and $\mu_m \in \R^d, \Sigma_m \in \R^{d\times d}$. Further, for the inference SDE we consider the Ornstein-Uhlenbeck process
\begin{equation}
    \mathrm d X_t = - AX_t \mathrm  dt + B \, \mathrm dW_t, \qquad X_0 \sim p_\mathrm{data},
\end{equation}
with matrices $A, B \in \R^{d\times d}$. For simplicity, let us consider $A = \alpha \mathbbm{1}$ and $B = \beta \mathbbm{1}$ with $\alpha, \beta \in \R$. Conditioned on an initial condition $x_0$, the marginal densities of $X$ are then given by
\begin{equation}
    p_t^\mathrm{SDE}(x; x_0) = \mathcal{N}\left(x; x_0 e^{-\alpha t}, \frac{\beta^2}{2\alpha}\left(1-e^{-2\alpha t} \right)\mathbbm{1} \right).
\end{equation}
We can therefore compute
\begin{subequations}
\begin{align}
    p^\mathrm{SDE}_t(x) &= \int_{\R^d} p^\mathrm{SDE}_t(x; x_0)p_\mathrm{data}(x_0) \mathrm d x_0 \\
    &= \sum_{m=1}^M \gamma_m \int_{\R^d}  \mathcal{N}\left(x; y_0 e^{-\alpha t}, \frac{\beta^2}{2\alpha}\left(1-e^{-2\alpha t}  \right)\mathbbm{1} \right)  \mathcal{N}(x_0; \mu_m, \Sigma_m) \mathrm d x_0 \\
    &= \sum_{m=1}^M \gamma_m  \mathcal{N}\left(x; e^{-\alpha t} \mu_m, \frac{\beta^2}{2\alpha}\left(1-e^{-2\alpha t}  \right)\mathbbm{1}+ e^{-2\alpha t} \Sigma_m  \right).
\end{align}
\end{subequations}
We can now readily compute the score $\nabla \log p_t^\mathrm{SDE}(x)$.

\subsection{Connecting the Ehrenfest process to score-based generative modeling}
\label{app: Ehrenfest - score based SDE}

As we have outline in \Cref{sec: Ehrenfest and score}, we can directly link the Ehrenfest process to score-based generative modeling in continuous time and space. In particular, we can use any model that has been trained in the typically used setting for our state-discrete Ehrenfest process. For instance, we can rely on DDPM models, which typically consider the forward SDE
\begin{equation}
\label{eq: DDPM forward process}
    \mathrm d X_t = -\frac{1}{2} \beta(t) X_t + \sqrt{\beta(t)} \mathrm dW_t
\end{equation}
on the time interval $[0, 1]$, where $\beta:[0, 1] \to \R$ is a function that scales time. This can be see by looking at the Fokker-Planck equation. For the process \eqref{eq: DDPM forward process} conditioned at the initial value $X_0 = x_0$ it holds 
\begin{equation}
    X_t \sim \mathcal{N}\left(\exp\left( -\frac{1}{2} \int_0^t \beta(s) \mathrm ds \right) x_0, 1 - \exp\left(-\int_0^t \beta(s) \mathrm ds \right) \right).
    \label{eq: ddpm forward}
\end{equation}
In practice, we choose $\beta(t) := \beta_\mathrm{min} + t (\beta_\mathrm{max} - \beta_\mathrm{min})$ with $\beta_\mathrm{min} = 0.1, \beta_\mathrm{max} = 20$, as suggested in \citet{song2020score}. Note that this typically guarantees that $X_1$ is approximately distributed according to $\mathcal{N}(0, 1)$, independent of $x_0$. For our experiments we use a model provided by the \href{https://huggingface.co/docs/diffusers/api/models/unet2d}{Diffuser package}. Note that this model is actually not the score, but the scaled score $\widetilde{s}$ and one needs the transformation
\begin{equation}
    \nabla \log p_t^\mathrm{SDE}(x) = -\frac{\widetilde{s}(x, 1000 \,t)}{\sqrt{1 - \exp\left(-\int_0^t \beta(s) \mathrm ds\right)}}.
\end{equation}

The DDPM framework implicitly trains a model $\widetilde{\varphi}(x, t)$ on
\begin{align}
    \mathcal{L}_\mathrm{OU}(\widetilde{\varphi}) &= \E\left[\left(\widetilde{\varphi}(x, t) - \nabla \log p_{t|0}^\mathrm{OU}(x|x_0)\right)^2\right] =  \E\left[\left(\widetilde{\varphi}(x, t) +\frac{\left(X_t-\mu_t(x_0) \right)}{\sigma_t^2} \right)^2\right],
\end{align}
which in an alternative formulation simplifies to predicting the noise $\varepsilon_t$ of
\begin{align}
    X_t = \underbrace{x_0 \exp\left( -\frac{1}{2} \int_0^t \beta(s) \mathrm ds \right)}_{\mu_t(x_0)}  + \varepsilon_t \underbrace{\sqrt{1 - \exp\left(-\int_0^t \beta(s) \mathrm ds\right)}}_{\sigma_t}  .
\end{align}
Since the terms $\mu_t(x_0)$ and $\sigma_t$ cancel, we arrive at the simplified loss
\begin{align}
    \mathcal{L}_\mathrm{OU}(\widetilde{\varphi}) 
    &= \E\left[\left(\widetilde{\varphi}(x, t) - \nabla \log p_{t|0}^\mathrm{OU}(x|x_0)\right)^2\right] =  \E\left[\left(\widetilde{\varphi}(x, t) +\frac{\varepsilon_t}{\sigma_t} \right)^2\right].
\end{align}

The Orstein-Uhlenbeck forward rates can be thus substituted by
\begin{equation}
    \E_{x_0 \sim p_{0|t}(x_0 | x)}\left[ \frac{p_{t|0}(x \pm \frac{2}{\sqrt{S}} | x_0)}{p_{t|0}(x | x_0)} \right] \approx 1 \pm \frac{2}{\sqrt{S}} \nabla \log p_{t}^\mathrm{OU}(x) \\
    = 1 \pm \frac{2}{\sqrt{S}} \widetilde{\varphi}(x, t).
\end{equation}

Similarly, the Taylor rates can be computed with 
\begin{align}
    \mathbb{E}_{x_0}\left[\frac{p_{t|0}\left(x \pm \delta \Big| x_0\right)}{p_{t|0}(x|x_0)}\right] 
    &\approx \exp\left( -\frac{\delta^2}{2 \sigma_t^2}\right) \left(1 \mp \frac{(x-\E_{x_0}\left[\mu_t(x_0)\right])\delta}{\sigma^2}\right) \\
    &= \exp\left( -\frac{\delta^2}{2 \sigma_t^2}\right) \left(1 \mp \frac{\widetilde{\varphi}(x, t)\delta}{\sigma_t}\right).
\end{align}

\subsection{Illustrative example}
\label{app: illustrative example}

As an illustrative example we choose a distribution which is tractable and perceivable.
We model a two dimensional distribution of pixels, which are distributed proportionally to the pixel value of an image of a capital ``E''. The visualization of the data distribution is governed by its $33 \times 33$ pixels and a single sample from the distribution is a black pixel indexed by its location $(x, y)$ on the $33 \times 33$ pixel grid. The diffusive forward process acts upon the coordinate and diffuses with progressing time the black pixels into a two dimensional (approximately) binomial distribution at time $t=1$.

We use the identical architecture as \cite{campbell2022continuous}, used for their illustrative example.
Subsequently, the architecture incorporates two residual blocks, each comprising a Multilayer Perceptron (MLP) with a single hidden layer characterized by a dimensionality of 32, a residual connection that links back to the MLP's input, a layer normalization mechanism, and ultimately, a Feature-wise Linear Modulation (FiLM) layer, which is modulated in accordance to the time embedding. The architecture culminates in a terminal linear layer, delivering an output dimensionality of $2$. 
The time embedding is accomplished utilizing the Transformer's sinusoidal position embedding technique, resulting in an embedding of dimension 32. This embedding is subsequently refined through an MLP featuring a single hidden layer of dimension 32 and an output dimensionality of 128. In order to generate the FiLM parameters within each residual block, the time embedding undergoes processing via a linear layer, yielding an output dimension of 2.

We test our proposed reverse rate estimators by training them to reconstruct the data distribution at time $t=0$. For evaluation, we draw $500.000$ individual pixels proportionally to the approximated equilibrium distribution and plot their respective histograms at time $t=0$ in Figure \ref{fig: toy example E}.

For training, we sample 1.000.000 pixel values proportional to the gray scale value of the 'E' image serving as the true data distribution.
We perform optimization with Adam with a learning rate of $0.001$ and optimize for 100.000 time steps with a batch size of 2.000.

\subsection{MNIST}
\label{app: MNIST}

The MNIST experiments were conducted with the scaled Ehrenfest process.
The MNIST data set consists of $28 \times 28$ gray scale images which we resized to $32 \times 32$ in order to be processable by use our standard DDPM architecture
We used $S=256^2$ states to ensure 256 states in the range of $[-1, 1]$ with a difference between states of of $\frac{2}{\sqrt{S}}$.
For optimization, we resorted to the default hyperparameters of Adam \cite{kingma2014adam} and used an EMA of 0.99 with a batch size of 128.
For the rates we chose the continuous DDPM schedule proposed by Song and we stopped the reverse process at $t=0.01$ due to vanishing diffusion and resulting high variance rates close to the data distribution.

\subsection{Image modeling with CIFAR-10}
\label{app: CIFAR}

We employ the standard DDPM architecture from \citet{ho2020denoising} and adapt the output layer to twice the size when required by the conditional expectation and the Gaussian predictor. 
The score and first order Taylor approximations did not need to be adapted.
For the ratio case, we adapted the architecture by doubling the final convolutional layer to six channels such that the first half (three channels) predicted the death rate and the second three channels predicted the birth rate.
For the time dependent rate $\lambda_t$ we tried the cosine schedule of \cite{nichol2021improved} and the variance preserving SDE schedule of \cite{song2020score}. The cosine schedule ensures the expected value of the scaled Ehrenfest process to converges to zero with $\mathbb{E}_{x_0}[x_t] = \cos\left(\frac{\pi}{2} \ t\right)^2 x_0$ and translates to a time dependent jump process rate of $\lambda_t = \frac{1}{4} \,\pi \tan\left(\frac{\pi}{2} \ t\right)$, which is unbounded close to the equilibrium distribution and therefore has to be clamped. We choose $\lambda_t \in [0, 500]$ in our case.
Due to numerical considerations regarding the exploding rates due to diminishing diffusion close to $t=0$, we restricted the reverse process to times $t\in [0.01, 1]$. In general, we can transform any deliberately long sampling time $T$ to $T=1$ via the time transformation of the master equation in \ref{app: time transformation}.

We use the standard procedure for training image generating diffusion models \cite{loshchilov2016sgdr}. 
In particular, we employ a linear learning rate warm up for $5.000$ steps and a cosine annealing from $0.0002$ to $0.00001$ with the Adam optimizer. The batch size was chosen as $256$ and an EMA with the factor $0.9999$ was applied for the model used for sampling. For sampling we ran the reverse process for $1.000$ steps and employed $\tau$-Leaping as showcased in \citet{campbell2022continuous} with a resulting $\tau=0.001$. We also utilized the predictor-corrector sampling method starting at $t=0.1$ to the minimum time of $t=0.01$. Whereas \citet{campbell2022continuous} reported significant gains performing corrector sampling, we observe behavior close to other state-continuous diffusion models which only apply few or no corrector steps at all.

\begin{figure}[ht]
    \centering
    \begin{minipage}{0.48\textwidth}
        \centering
        \includegraphics[width=\linewidth]{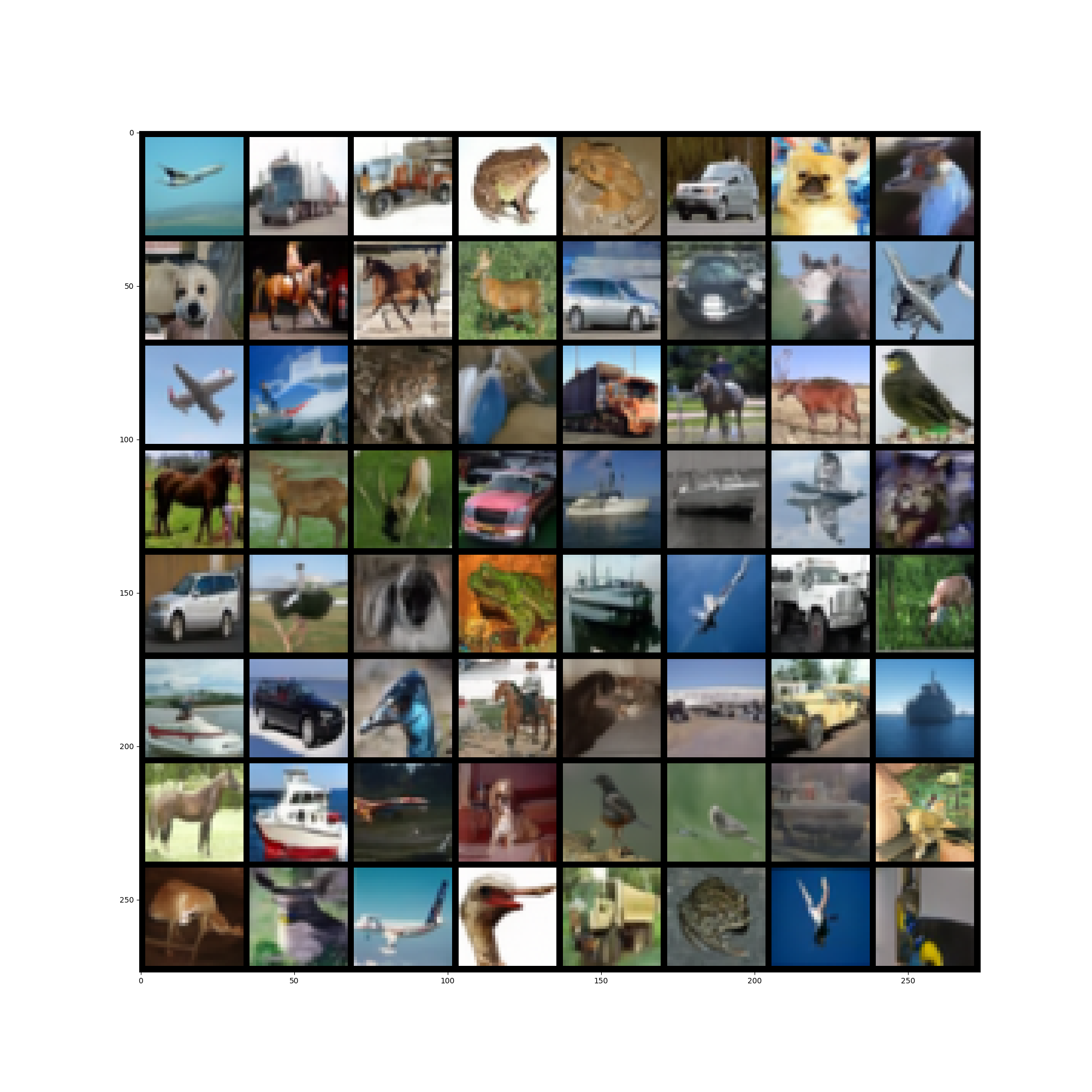}
        \caption{Samples from the reverse scaled Ehrenfest process obtained by finetuning the DDPM architecture with $\mathcal{L}_\text{Taylor}$ (\ref{eq: first Taylor ratio loss}).}
        \label{fig:cifar 10 taylor big 1}
    \end{minipage}\hfill
    \begin{minipage}{0.48\textwidth}
        \centering
        \includegraphics[width=\linewidth]{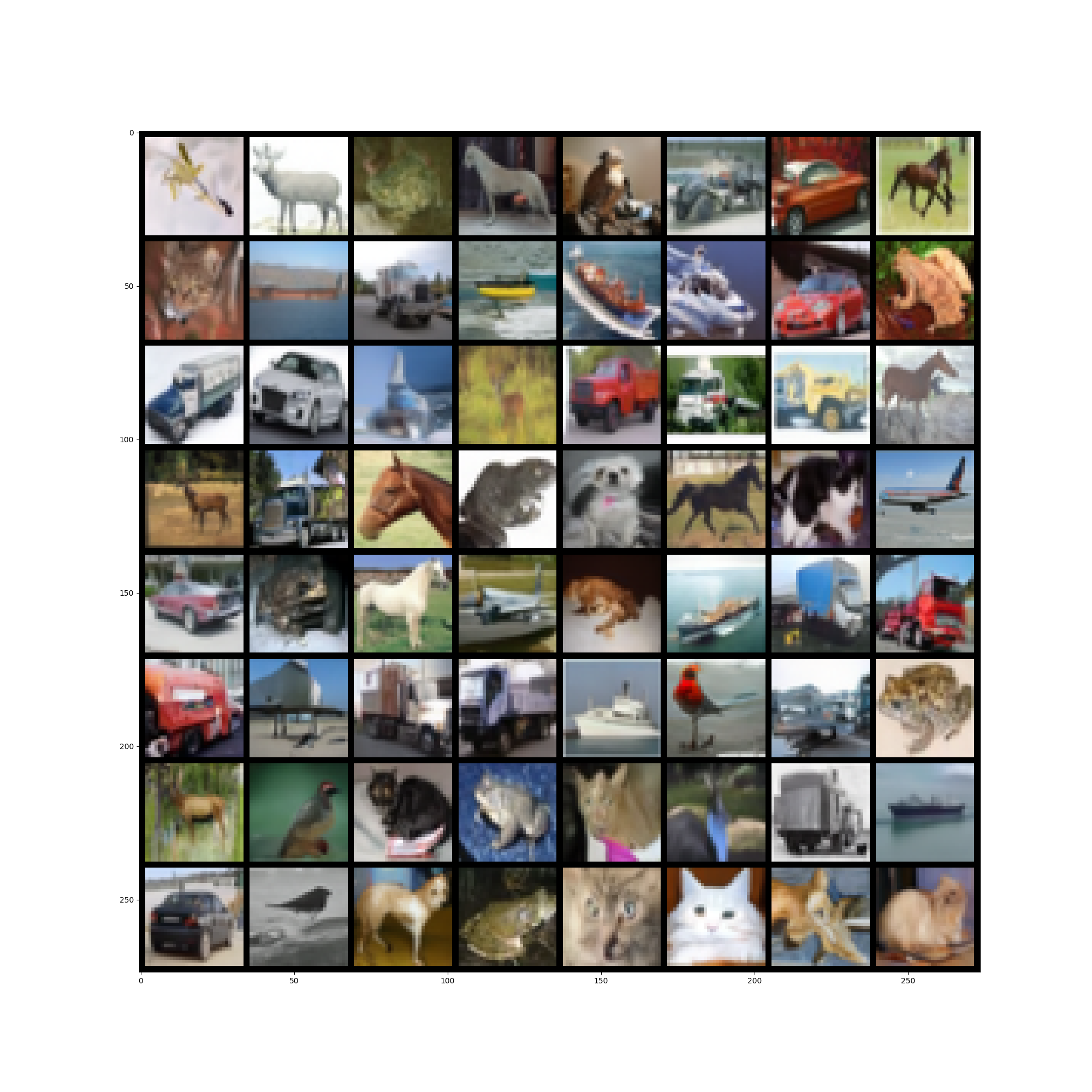} 
        \caption{Samples from the reverse scaled Ehrenfest process obtained by finetuning the DDPM architecture with $\mathcal{L}_\text{Taylor}$ (\ref{eq: first Taylor ratio loss}).}
        \label{fig:cifar 10 taylor big 2}
    \end{minipage}
\end{figure}

\begin{figure}[ht]
    \centering
    \begin{minipage}{0.48\textwidth}
        \centering
        \includegraphics[width=\linewidth]{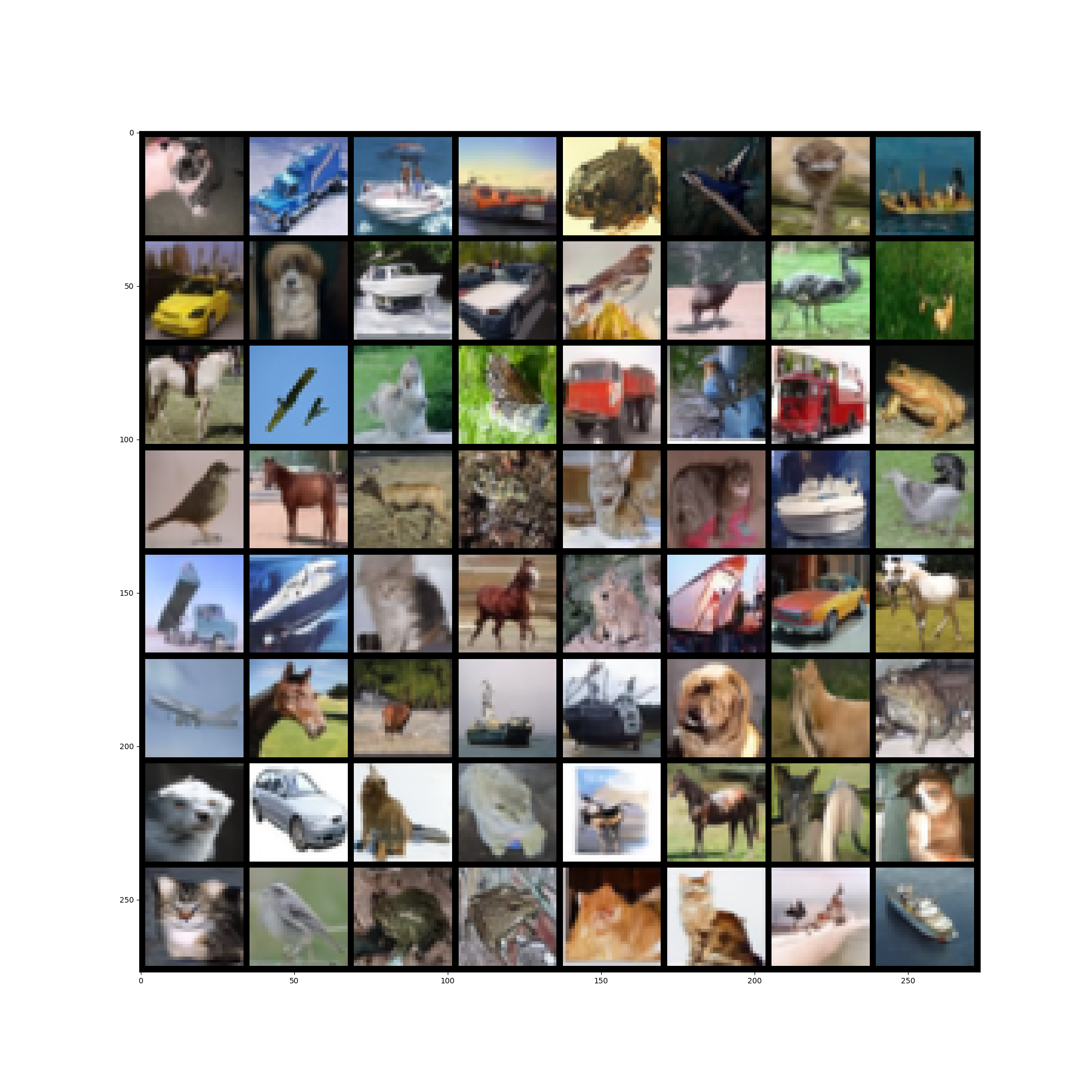} 
        \caption{Samples from the reverse scaled Ehrenfest process obtained by finetuning the DDPM architecture with $\mathcal{L}_\text{OU}$ (\ref{eq: forward OU loss}).}
        \label{fig:cifar 10 score big 1}
    \end{minipage}\hfill
    \begin{minipage}{0.48\textwidth}
        \centering
        \includegraphics[width=\linewidth]{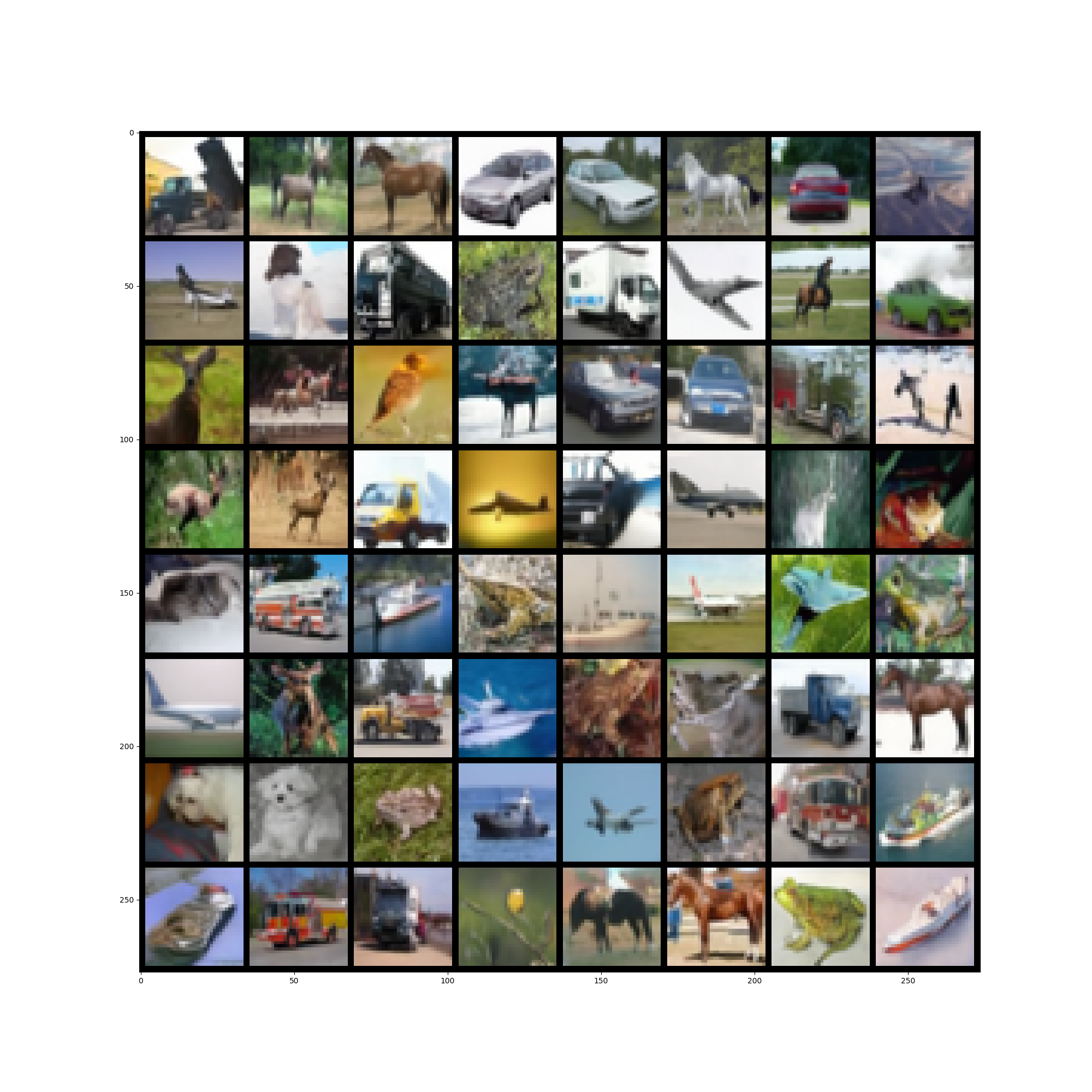} 
        \caption{Samples from the reverse scaled Ehrenfest process obtained by finetuning the DDPM architecture with $\mathcal{L}_\text{OU}$ (\ref{eq: forward OU loss}).}
        \label{fig:cifar 10 score big 2}
    \end{minipage}
\end{figure}

\end{document}